\documentclass[11pt]{article}

\usepackage{acl}

\usepackage{times}
\usepackage{latexsym}
\usepackage[T1]{fontenc}
\usepackage[utf8]{inputenc}
\usepackage{microtype}
\usepackage{inconsolata}
\usepackage{graphicx}
\usepackage{xcolor}

\usepackage{amsmath}
\usepackage{amssymb}
\usepackage{bm}
\usepackage{mathtools}
\usepackage{algorithm}
\usepackage{algpseudocode}
\usepackage{enumitem}
\usepackage{booktabs}
\usepackage{multirow}
\usepackage{subcaption}
\usepackage{placeins}
\usepackage[breakable,skins]{tcolorbox}

\hypersetup{colorlinks=true,linkcolor=blue,urlcolor=blue,citecolor=blue}

\newtcolorbox{promptbox}[2][]{%
  enhanced,
  breakable,
  colback=blue!4,
  colframe=blue!45,
  boxrule=0.6pt,
  arc=3pt,
  left=8pt, right=8pt, top=5pt, bottom=5pt,
  fonttitle=\bfseries\small\sffamily,
  coltitle=blue!55!black,
  colbacktitle=blue!12,
  title={#2},
  #1
}

\newcommand{\Rplan}{R_{\mathrm{plan}}}
\newcommand{\Rout}{R_{\mathrm{out}}}
\newcommand{\Jpsi}{J_{\psi}}
\newcommand{\Rwarm}{R_{\mathrm{warm}}}
\newcommand{\Lsolve}{\mathcal{L}_{\mathrm{solve}}}
\newcommand{\Lplan}{\mathcal{L}_{\mathrm{plan}}}
\newcommand{\LRM}{\mathcal{L}_{\mathrm{RM}}}
\newcommand{\Pass}{\text{Pass}}

\newcommand{\ind}{\mathbb{I}}
\newcommand{\zhat}{\hat{z}}
\newcommand{\Ksolve}{K_{\mathrm{solve}}}
\newcommand{\Ktuple}{K_{\mathrm{tuple}}}
\newcommand{\std}[1]{$_{\pm\,#1}$}

\title{Cast a Wider Net: Coordinated Pass@K Policy Optimization for Code Reasoning}

\author{
  Yilong Li and Suman Banerjee \\
  University of Wisconsin--Madison \\
  \texttt{\{yli758,suman\}@wisc.edu}
  \And
  Tong Che\thanks{Project Lead, Corresponding Author.} \\
  NVIDIA Research \\
  \texttt{tongc@nvidia.com}
}

\begin{document}
\maketitle

\begin{abstract}
While repeated sampling against a verifier is the standard test-time compute strategy for code generation (measured by pass@$K$), drawing $K$ independent samples from a single distribution often wastes the compute budget on redundant, near-duplicate reasoning paths.
This failure is costly in competitive programming, where many problems
admit multiple distinct algorithmic strategies and passing pass@$K$
requires only one correct attempt. We propose Coordinated Pass@$K$ Policy Optimization (CPPO), which
trains a joint planner--solver policy for pass@$K$: a planner emits
a tuple of $K{=}4$ alternative high-level methods, and a shared solver attempts one
solution per method. We train it with a multiplicative planner
reward, $R_{\mathrm{plan}} = J_\psi \cdot R_{\mathrm{out}}$, assigning credit only
to valid strategy tuples that lead to verifier-confirmed pass@$K$ success. Across
APPS, CodeContests, and LiveCodeBench-v6, CPPO improves pass@$4$ over direct
sampling, planning baselines, planner-only SFT, and pass@$K$-oriented RL under
the same $K{=}4$ solver-attempt budget, with statistically significant gains on
six of nine model--benchmark cells; for instance, it lifts
Qwen3.5-9B LiveCodeBench-v6 from $0.588$ (PKPO, the strongest baseline)
to $0.728$ ($+0.14$; hierarchical bootstrap, $p<0.05$).
\end{abstract}

\section{Introduction}
\label{sec:intro}
Repeated sampling against a verifier is the standard way to allocate test-time compute in code generation~\citep{brown2024monkeys,snell2024scaling}, with pass@$K$ as the canonical metric for competitive-programming benchmarks~\citep{chen2021codex,li2022competition,jain2024livecodebench}. Given a fixed attempt budget, the central question is how to allocate attempts effectively.

Standard pipelines spend this budget on $K$ independent samples from one answer
distribution. Once the model is moderately confident, these samples tend to
collapse onto the dominant mode: on a programming problem, this can mean several
rollouts that implement the same algorithm with only cosmetic changes---for
example, four memoized variants of one dynamic-programming solution---even when
the problem admits distinct alternatives. Recent pass@$K$ training methods such
as pass@$K$-only RL~\citep{chen2025passk} and
PKPO~\citep{walder2025pkpo}, as well as diversity-aware RL methods such as
Darling~\citep{li2025darling}, improve how answer samples are rewarded, but they
do not model the $K$ attempts as one coordinated action: the attempts remain
separate draws from the answer distribution, so the same mode-dropping failure
can persist.

\begin{figure*}[t]
\centering
\includegraphics[width=0.85\linewidth]{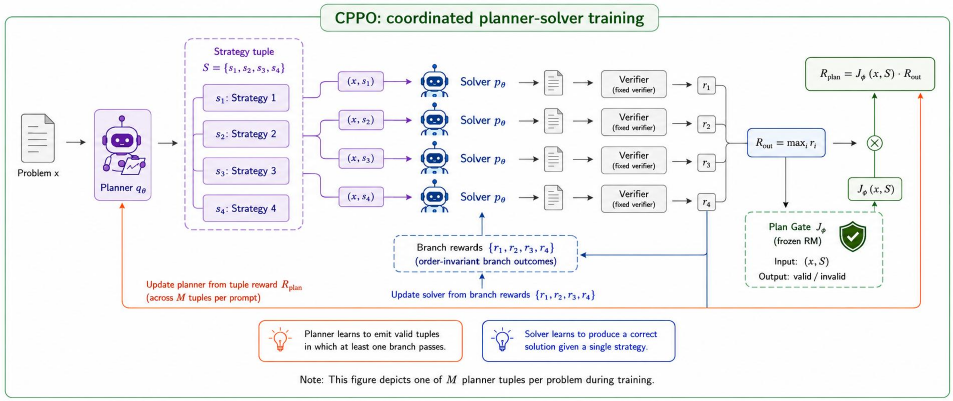}
\caption{Overview of Coordinated Pass@$K$ Policy Optimization. The planner $q_\Theta$ emits a strategy tuple $S=(s_1,\ldots,s_K)$; the shared solver $p_\Theta$ produces one solution per strategy; a verifier returns per-branch outcomes $r_i\in\{0,1\}$, and the outcome reward $\Rout=\max_i r_i$ scores pass@$K$ success. A frozen reward model $\Jpsi(x,S)$ gates plan validity, giving the planner reward $\Rplan=\Jpsi(x,S)\cdot\Rout$, which is nonzero only when the tuple is accepted by $\Jpsi$ \emph{and} at least one of its solver branches passes. Solver tokens are updated from within-tuple outcome advantages; planner tokens are updated from across-tuple advantages of $\Rplan$.}
\label{fig:arch}
\end{figure*}

Rather than relying on independent draws, we train the model to generate the $K$ attempts jointly. We focus on code reasoning, where competitive-programming
problems often admit multiple algorithmic strategies and any one correct
attempt is enough to pass. A planner jointly emits a structured
\emph{strategy tuple} of $K{=}4$ high-level methods, each conditioned on
earlier ones, and a shared solver attempts one solution per method. Improving
pass@$K$ therefore becomes a problem of producing complementary alternatives
rather than resampling from a single answer distribution. To our knowledge,
CPPO is the first RLVR post-training method that optimizes an autoregressive
strategy-tuple planner under a verifier-confirmed pass@$K$ objective. We call
this method Coordinated Pass@$K$ Policy Optimization (CPPO;
Figure~\ref{fig:arch}); the full training algorithm is in
\S\ref{sec:pipeline}.\footnote{Here, CPPO denotes \emph{Coordinated}
Pass@$K$ Policy Optimization: a joint planner--solver policy trained with the
multiplicative planner reward $\Rplan=\Jpsi\cdot\Rout$.}

\begin{figure}[htb]
\centering
\includegraphics[width=0.9\linewidth]{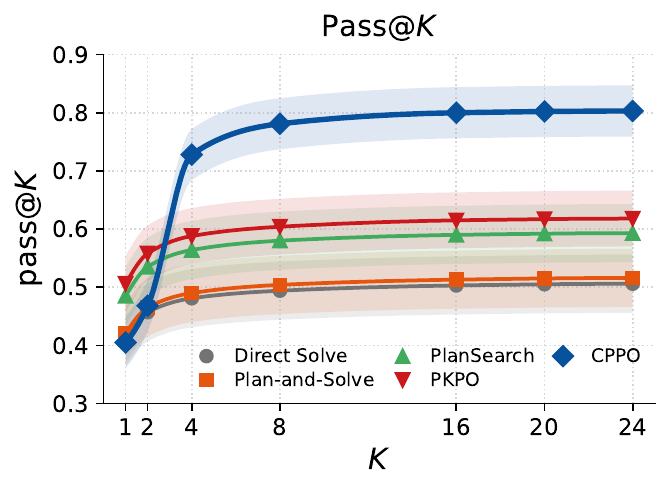}
\vspace{-0.4ex}
\caption{LiveCodeBench-v6 pass@$K$ for Qwen3.5-9B (4B counterpart in
Figure~\ref{fig:passk-delta-full}); all CPPO points use one
$\Ktuple{=}4$ planner and pool $4$-tuple rollouts for $K{>}4$
(Table~\ref{tab:k-sweep-decomposition}).}
\vspace{-0.5ex}
\label{fig:passk-delta}
\end{figure}

Table~\ref{tab:main} evaluates CPPO against top baselines under a fixed $K{=}4$ budget. Full CPPO consistently outperforms Tuple Planner SFT—yielding pass@$4$ gains of $+0.07$ to $+0.24$—which clearly isolates the value of joint planner--solver RL over SFT initialization alone.
The contributions can be summarized:
\begin{itemize}[leftmargin=*,itemsep=2pt,topsep=2pt]
\item \textbf{Verifier-trained strategy tuples for pass@$K$.}
We recast pass@$K$ RLVR from independently scored answer samples to an
autoregressive planner that emits a joint strategy tuple, separating CPPO from
prior pass@$K$-only RL~\citep{chen2025passk} and
PKPO~\citep{walder2025pkpo}.

\item \textbf{Multiplicative planner reward.} We introduce a multiplicative planner reward,
\[\Rplan = \Jpsi \cdot \Rout,
\]
that assigns credit to the planner only if the learned validity gate $\Jpsi$ accepts the strategy tuple and at least one solver branch succeeds. This gate acts as a strict validity constraint, primarily suppressing credit for malformed or duplicate plans (\S\ref{sec:component-ablation}), whereas verifier-confirmed solver success ($\Rout$) provides the pass@$K$ signal. We optimize this objective end-to-end using split-region GRPO updates for both planner and solver tokens (\S\ref{sec:objective}).

\item \textbf{Pass@$4$ gains at fixed budget.}
Across APPS, CodeContests, and LiveCodeBench-v6, CPPO improves
pass@$4$ over the baseline in all nine (size, benchmark)
cells of Table~\ref{tab:main} and Appendix~\ref{app:main-full}
under the same $K{=}4$ solver-attempt budget and verifier,
with six cells reaching $p < 0.05$ under a hierarchical
problem-and-seed bootstrap; on
Qwen3.5-9B LiveCodeBench-v6 the gain over PKPO is $+0.14$
($0.588 \rightarrow 0.728$), and the gains persist at larger
attempt budgets (Fig.~\ref{fig:passk-delta}).
\end{itemize}

% Main-results teaser table (page 1). TODO results: Table 2 (reasoning),
% Table 3 (ablations: -Jpsi gate / -outcome / fixed-K vs honest-M / warm-up),
% and the pass@k curve figure.
\begin{table*}[t]
\centering
\caption{Pass@$4$ results for Qwen3.5-4B and 9B, evaluated with a shared verifier and a constant $K{=}4$ solver-attempt budget.}
\label{tab:main}
\scriptsize
\setlength{\tabcolsep}{3pt}
\renewcommand{\arraystretch}{0.85}
\begin{tabular}{@{}llccc@{}}
\toprule
Method & Base model & APPS~\citep{hendrycks2021apps} & \shortstack{CodeContests~\citep{li2022competition}} & \shortstack{LiveCodeBench~(LCBv6)~\citep{jain2024livecodebench}} \\
\midrule
Direct Solve & Qwen3.5-4B & 0.515\std{0.022} & 0.094\std{0.014} & 0.214\std{0.013} \\
Direct Solve & Qwen3.5-9B & 0.696\std{0.021} & 0.175\std{0.019} & 0.481\std{0.014} \\
Plan-and-Solve & Qwen3.5-4B & 0.530\std{0.019} & 0.098\std{0.016} & 0.255\std{0.012} \\
Plan-and-Solve & Qwen3.5-9B & 0.801\std{0.015} & 0.171\std{0.017} & 0.490\std{0.020} \\
PlanSearch & Qwen3.5-4B & 0.554\std{0.022} & 0.128\std{0.021} & 0.236\std{0.012} \\
PlanSearch & Qwen3.5-9B & 0.770\std{0.019} & 0.168\std{0.018} & 0.564\std{0.019} \\
Pass@$K$ Training / RLVR & Qwen3.5-4B & 0.607\std{0.017} & 0.115\std{0.020} & 0.316\std{0.019} \\
Pass@$K$ Training / RLVR & Qwen3.5-9B & 0.762\std{0.015} & 0.185\std{0.021} & 0.503\std{0.020} \\
PKPO & Qwen3.5-4B & 0.722\std{0.021} & 0.156\std{0.020} & 0.488\std{0.015} \\
PKPO & Qwen3.5-9B & 0.787\std{0.020} & 0.183\std{0.025} & 0.588\std{0.015} \\
UpSkill & Qwen3.5-4B & 0.661\std{0.023} & 0.121\std{0.019} & 0.411\std{0.016} \\
UpSkill & Qwen3.5-9B & 0.762\std{0.015} & 0.136\std{0.017} & 0.542\std{0.013} \\
\midrule
Tuple Planner SFT & Qwen3.5-4B & 0.548\std{0.024} & 0.147\std{0.021} & 0.348\std{0.018} \\
Tuple Planner SFT & Qwen3.5-9B & 0.733\std{0.016} & 0.279\std{0.026} & 0.514\std{0.015} \\
\textbf{CPPO} & Qwen3.5-4B & 0.784$^{\dagger}$\std{0.013} & 0.231$^{\dagger}$\std{0.025} & 0.505\std{0.017} \\
\textbf{CPPO} & Qwen3.5-9B & 0.806\std{0.019} & 0.396$^{\dagger}$\std{0.028} & 0.728$^{\dagger}$\std{0.016} \\
\bottomrule
\end{tabular}

\smallskip
\begin{minipage}{\textwidth}
\scriptsize
\textit{Tuple Planner SFT} emits all $K$ plans in one autoregressive tuple,
$q_\Theta(s_i\mid x,s_{<i})$; CPPO further trains this planner--solver policy
with validity-gated outcome rewards. The matched iid-inference ablation is in
Appendix~\ref{app:inference-mode}.

\smallskip
$^{\dagger}$ CPPO significantly outperforms the strongest non-CPPO baseline
in that cell under a hierarchical bootstrap over problems and training seeds
($p<0.05$, $1000$ resamples; per-seed CIs in Appendix~\ref{app:seed-significance}).
Unmarked CPPO gains are positive but not significant.

\smallskip
\textit{Baselines.} We compare independent sampling, prompt-based planning,
inference-time plan search, pass@$K$-oriented RL, diversity-oriented RL, and
planner-only SFT; details are in \S\ref{sec:evaluation} and
Appendix~\ref{app:baselines}.
\end{minipage}
\end{table*}

\section{Background and Related Work}
\label{sec:background}

\paragraph{RLVR and optimizer.}
We work in the Reinforcement Learning with Verifiable Rewards (RLVR) setting popularized by DeepSeek-R1~\citep{guo2025deepseekr1}: a rule-based verifier $V(x, y) \in \{0, 1\}$ returns whether response $y$ is correct for prompt $x$, and this binary signal serves as the reward for policy optimization. We use GRPO~\citep{shao2024deepseekmath} with group-normalized advantages and a clipped, KL-regularized objective~\citep{schulman2017ppo}. CPPO keeps this optimizer fixed and changes only the policy factorization, reward, and token regions to which advantages are applied; Appendix~\ref{app:grpo-details} gives the standard GRPO equations.

\paragraph{Pass@$K$ objective.}
With a fixed attempt budget $K$ per prompt, pass@$K$ is the indicator that at least one of $K$ attempts is correct,
\begin{equation}
R_{\text{pass@}K}(x, y_{1:K}) \;=\;
\ind\!\left\{\max_{i=1}^{K} V(x, y_i) = 1\right\}.
\label{eq:passatk}
\end{equation}
The standard repeated-sampling factorization draws the $K$ attempts independently from a single answer distribution $p_\Theta(\cdot \mid x)$,
\begin{equation}
y_i \sim p_\Theta(\cdot \mid x), \quad i = 1, \ldots, K.
\label{eq:repeated-sampling}
\end{equation}

\paragraph{Mode coverage under independent sampling.}
A simple bound illustrates the limitation of drawing attempts
independently. If $p_\Theta(\cdot \mid x)$ places probability $\epsilon$ on a
correct-answer region $\mathcal{A}$, then none of $K$ iid samples lands in
$\mathcal{A}$ with probability $(1-\epsilon)^K$; for $\epsilon{=}0.05$ even
$K{=}16$ leaves a missing-mass probability of $\approx 0.44$. Re-weighting the
per-sample reward (pass@$K$-only RL, PKPO) reshapes $p_\Theta$---and can raise
$\epsilon$---but the $K$ draws stay independent, with no explicit negative
dependence or branch-level allocation across strategy regions. CPPO instead
introduces a structured joint action: its $K$ branches are sampled jointly
through the planner (\S\ref{sec:design}) and can be allocated to complementary
strategies within a single trajectory.

\paragraph{Repeated sampling and test-time compute.}
Repeated sampling is the standard way to allocate test-time compute for code generation~\citep{brown2024monkeys}, and compute-optimal sampling can rival substantially larger models~\citep{snell2024scaling}; CPPO targets this same pass@$K$ regime but changes how the $K$ attempts are generated.\footnote{Results in this version use an updated training-data setup for the planner (\S\ref{sec:train-data}); the relative ordering of methods in Table~\ref{tab:main} and our conclusions are unchanged.}

\paragraph{Pass@$K$ training and diversity rewards.}
Recent post-training methods improve repeated sampling by changing how answer samples receive credit. Pass@$K$ training~\citep{chen2025passk} and PKPO~\citep{walder2025pkpo} optimize set-level pass@$K$ rewards over sampled answers, while Darling~\citep{li2025darling}, DIVER~\citep{hu2025diver}, and UpSkill~\citep{shah2026upskill} add diversity pressure through reward shaping, exploration bonuses, or latent-skill regularization. These methods can broaden the answer distribution, but they do not model the $K$ attempts as a single coordinated policy action. CPPO instead treats the $K$-way set as a structured action: one planner rollout emits a coordinated tuple of strategies, and the solver executes one branch per strategy.

\paragraph{Verifiers and LLM-as-judge.}
Outcome verifiers~\citep{cobbe2021verifiers} and process reward models~\citep{lightman2023prm} score model outputs without ground-truth labels. CPPO's plan-validity reward model is a small generative verifier in this tradition, with labels obtained via LLM-as-judge~\citep{zheng2023judge}.

\paragraph{Planning for reasoning.}
Planning methods add structure before generation rather than after scoring. Plan-and-solve prompting~\citep{wang2023plan}, Tree of Thoughts~\citep{yao2023tot}, and PlanSearch~\citep{wang2024plansearch} use natural-language plans or search procedures at inference time over a frozen model. They show that explicit plans can help reasoning, but they do not train a planner under the pass@$K$ objective. CPPO brings planning into post-training: the planner learns to emit a single autoregressive strategy tuple whose later items condition on earlier ones, and receives validity-gated credit only when the tuple leads to verifier-confirmed pass@$K$ success.

\section{Coordinated Pass@K Policy Optimization}
\label{sec:methodology}

CPPO changes the policy factorization: instead of drawing $K$ iid answers, it samples one tuple of $K{=}4$ algorithmic strategies. This factorization splits credit assignment: the solver must learn which strategy-conditioned branch succeeds, and the planner must learn which strategy tuple makes at least one branch succeed. CPPO therefore assigns branch-level outcome rewards to solver tokens and a tuple-level reward to planner tokens. Because tuple-level rewards are sparse and can be gamed by malformed plans, we gate planner credit with a narrow plan-validity reward model and train the system through a staged procedure. Table~\ref{tab:method-checks} contrasts CPPO with the closest planning and pass@$K$ post-training baselines along the design axes that follow.

\begin{table}[!htb]
\centering
\caption{Comparison of CPPO with the closest planning and pass@$K$
post-training baselines along four axes: output granularity, search
strategy, intervention stage, and RLVR training.}
\label{tab:method-checks}
\scriptsize
\setlength{\tabcolsep}{3pt}
\renewcommand{\arraystretch}{0.87}
\begin{tabular}{@{}lcccc@{}}
\toprule
Method & Granularity & \shortstack{Search\\strategy} & Stage & \shortstack{RLVR\\ready} \\
\midrule
Plan-and-Solve       & answer      & iid                & inference     & --         \\
PlanSearch           & plan features & enumerate          & inference     & --         \\
Pass@$K$-only RL     & answer      & iid                & post-training & \checkmark \\
PKPO                 & answer      & iid                & post-training & \checkmark \\
\midrule
\textbf{CPPO (ours)} & \textbf{strategy} & \textbf{per-item} & \textbf{both} & \checkmark \\
\bottomrule
\end{tabular}
\end{table}

\subsection{A Joint Strategy-Tuple Policy}
\label{sec:design}

Standard pass@$K$ samples $K$ answers independently from one distribution. CPPO instead samples one structured object: a tuple of $K$ strategy sketches generated autoregressively, so each method conditions on its predecessors and avoids redundant branches. The planner emits the tuple
\begin{equation}
S = (s_1, \ldots, s_K),
\label{eq:tuple}
\end{equation}
with factorization
\begin{equation}
q_\Theta(S \mid x) = \prod_{i=1}^{K} q_\Theta(s_i \mid x, s_{<i}).
\label{eq:planner-policy}
\end{equation}
A shared solver produces one answer per method,
\begin{equation}
y_i \sim p_\Theta(\cdot \mid x, s_i), \quad i = 1, \ldots, K,
\label{eq:solver-policy}
\end{equation}
yielding the full trajectory
\begin{equation}
\tau = (S, y_{1:K})
\label{eq:trajectory}
\end{equation}
with probability
\begin{equation}
\pi_\Theta(\tau \mid x) = q_\Theta(S \mid x) \prod_{i=1}^{K} p_\Theta(y_i \mid x, s_i).
\label{eq:traj-prob}
\end{equation}
In the simplest implementation, a single backbone with shared parameters $\Theta$ induces both $q_\Theta$ (planner factor) and $p_\Theta$ (solver factor) from two prompt modes: the \textsc{plan} prompt produces a concise, task-focused strategy tuple (planner tokens), and the \textsc{solve} prompt produces one answer conditioned on a single strategy (solver tokens). Planner and solver losses then apply to these disjoint token regions (exact prompts in Appendix~\ref{app:prompts}).

\subsection{Coordinated Rewards and Credit Assignment}
\label{sec:reward-model}
\label{sec:objective}
\label{sec:outcome}
\label{sec:plan-validity-def}
\label{sec:failure-modes}
\label{sec:llm-judge-rm}
\label{sec:coordinated-reward}

CPPO assigns credit at two granularities. Solver tokens receive branch-level
verifier feedback: for branch \(i\), \(r^{\mathrm{out}}_i=V(x,y_i)\), and the
tuple-level outcome reward is the pass@$K$ indicator
\begin{equation}
\Rout(x, y_{1:K}) =
\ind\!\left\{\max_i V(x, y_i) = 1\right\}.
\label{eq:tuple-outcome}
\end{equation}
This outcome signal is appropriate for solver branches, but it does not tell the
planner whether the visible tuple is well formed, non-duplicative, or free of
answer leakage. We therefore gate planner credit with a narrow plan-validity
model \(\Jpsi(x,S)\). The gate checks that the tuple parses into \(K\) methods,
avoids literal duplicates, contains no final answer or completed implementation,
specifies actionable approaches, and stays on topic. It does not predict
whether a plan will solve the task. CPPO uses its binary decision
\begin{equation}
\Jpsi(x, S) =
\ind\!\left\{p_\psi(\Pass \mid x, S) \geq \tau\right\} \in \{0, 1\},
\label{eq:jpsi}
\end{equation}
with \(\tau\) chosen for high recall on validation data
(\S\ref{sec:models}). Since the gate learns only this minimal contract, a small
reward model suffices; training details are given in Stage~2 of
\S\ref{sec:pipeline}, and \S\ref{sec:validity-gate-diag} validates its use as a
gate rather than a quality scorer.
The planner reward combines validity and outcome:
\begin{equation}
\Rplan(x, S, y_{1:K}) = \Jpsi(x, S) \cdot \Rout(x, y_{1:K}).
\label{eq:plan-reward}
\end{equation}
Thus planner credit is given only to tuples that are valid and yield at least
one passing strategy-conditioned solution: \(\Rout\) rejects valid but
unsuccessful plans, while \(\Jpsi\) rejects successes obtained through malformed,
duplicated plans or plans that leak the answer.

\paragraph{Split-region advantages.}
\label{sec:solver-adv}
\label{sec:planner-adv}
\label{sec:loss}
\label{sec:algorithm}
The two token regions carry different credit granularity. Solver
advantages compare the $K$ branches within one tuple $\tau$,
\begin{equation}
a_i(\tau, x) = \frac{r^{\mathrm{out}}_i - \mathrm{mean}(r^{\mathrm{out}}_{1:K})}{\mathrm{std}(r^{\mathrm{out}}_{1:K}) + \epsilon},
\label{eq:solver-adv}
\end{equation}
and update answer tokens through $\log p_\Theta(y_i \mid x, s_i)$.
Planner advantages normalize $\Rplan$ across $M$ tuples sampled for
the same prompt,
\begin{equation}
A_{\mathrm{plan}}^{(m)} = \frac{\Rplan^{(m)} - \mathrm{mean}(\Rplan^{(1:M)})}{\mathrm{std}(\Rplan^{(1:M)}) + \epsilon},
\label{eq:planner-adv}
\end{equation}
and update planner tokens through $\log q_\Theta(S^{(m)} \mid x)$. The
two advantages enter the standard clipped, KL-regularized GRPO
objective of \eqref{eq:grpo-loss}:
\begin{equation}
\mathcal{L} = \sum_{m=1}^{M} \sum_{i=1}^{K} \Lsolve^{(m,i)}\!\bigl(a_i^{(m)}\bigr) + \sum_{m=1}^{M} \Lplan^{(m)}\!\bigl(A_{\mathrm{plan}}^{(m)}\bigr),
\label{eq:cppo-loss}
\end{equation}
where $\Lsolve$ and $\Lplan$ denote the negative GRPO objective on solver and
planner token spans. In the shared-backbone implementation of \S\ref{sec:design}, the
planner and solver losses apply to disjoint token spans (selected by
the \textsc{plan} and \textsc{solve} prompts) but update the same
parameters $\Theta$.

% Training-objective content has been folded into reward_model.tex (split-region advantages) and pipeline.tex (warm-up and joint loop).

\subsection{Making the Sparse Joint Reward Trainable}
\label{sec:pipeline}

The planner reward $\Rplan=\Jpsi\cdot\Rout$ is nonzero only when the planner emits a valid tuple and at least one solver branch passes. Since clean $K$-method tuples and successful rollouts are rare at initialization, CPPO first raises reward density through planner SFT, validity-gate training, RM-guided warm-up, and a reward-density audit.

\paragraph{Stage 1: Planner SFT.}
We initialize the planner by maximum likelihood on strict-$K$ gold tuples---self-generated plan samples filtered by a Qwen3.5-9B judge (Appendix~\ref{app:sft-funnel}). All experiments use the same $K{=}4$ planner; the $K$ sweep in \S\ref{sec:component-ablation} obtains smaller or larger budgets by truncating or pooling planner rollouts at inference time, without retraining. The CPPO reward is set-level, and the SFT data does not assign semantic roles to tuple positions; to confirm that truncating to the first $K{<}4$ outputs introduces no material positional artifact, an order-randomized control in Appendix~\ref{app:inference-mode} reports first-$K$, last-$K$, and random-$K$ selection within seed variance of each other.

\paragraph{Stage 2: Validity gate.}
\label{sec:rm-training}
We train $p_\psi$ on a balanced CodeContests tuple pool of
LLM-judged~\citep{zheng2023judge} strict-four positives and
semantically invalid but superficially well-formed negatives, and
accept a checkpoint only when pre-registered validation metrics pass.
During CPPO, $\Jpsi$ is frozen within each phase and refreshed between
phases on a fresh batch of planner outputs; exact splits, thresholds,
and the refresh schedule are in
Appendix~\ref{app:rm-training-details}.

\paragraph{Stage 3: RM-guided warm-up.}
\label{sec:warmup}
We precede joint training with a planner-only stage under
$\Rwarm(x, S) = \Jpsi(x, S)$, applied to planner tokens through the
GRPO objective of \eqref{eq:grpo-loss}, until the fraction of plans
accepted by $\Jpsi$ plateaus.

\paragraph{Stage 4: Reward-density audit.}
We run a forward-only evaluation pass and require a nonzero frozen-solver pass@$K$
rate from the warmed planner together with a nonzero $\Rplan$ density
across sampled rollouts; if either fails, we return to Stage~1 or
Stage~2 rather than launch CPPO.

\paragraph{Stage 5: Joint CPPO update.}
The inner loop of joint training is summarized in
Algorithm~\ref{alg:cppo-main}: $M$ planner tuples per prompt, $K$
branches each, within-tuple solver advantages and across-tuple planner
advantages feed the split-region GRPO loss
\eqref{eq:cppo-loss}. The full pipeline pseudocode is in
Appendix~\ref{app:algorithm}.

\begin{algorithm}[htb]
\caption{One CPPO update step (Stage~5).}
\label{alg:cppo-main}
\scriptsize
\begin{algorithmic}[1]
\Require prompt $x$, $M$ planner tuples per prompt, $K$ branches per tuple
\State Sample $S^{(m)} \sim q_\Theta(\cdot \mid x)$ for $m = 1, \ldots, M$.
\State Score validity $\Jpsi(x, S^{(m)})$ with the frozen RM.
\State Solve $y_i^{(m)} \sim p_\Theta(\cdot \mid x, s_i^{(m)})$ for every branch.
\State Verify $r_i^{(m)} = V(x, y_i^{(m)})$ and form $\Rout^{(m)}$, $\Rplan^{(m)} = \Jpsi(x, S^{(m)}) \cdot \Rout^{(m)}$.
\State Compute within-tuple solver advantages $a_i^{(m)}$ from \eqref{eq:solver-adv}.
\State Compute across-tuple planner advantages $A_{\mathrm{plan}}^{(m)}$ from \eqref{eq:planner-adv}.
\State Update $\Theta$ with the split-region GRPO loss \eqref{eq:cppo-loss}.
\end{algorithmic}
\end{algorithm}
\FloatBarrier

\section{Experimental Setup}
\label{sec:setup}

We instantiate CPPO on competitive programming, where the verifier $V$ executes each candidate program against the problem's tests inside a sandboxed, time-limited runner (\S\ref{sec:outcome}).

\subsection{Models}
\label{sec:models}

Unless otherwise stated, the planner and solver share the same backbone and are switched by the \textsc{plan} and \textsc{solve} prompt modes of \S\ref{sec:design}; the two token regions are trained jointly in full CPPO. The main experiments evaluate Qwen3.5-\{2B, 4B, 9B\}~\citep{qwen2026qwen35}; we additionally verify generality on Gemma~4~\citep{google2026gemma4} in \S\ref{sec:generality-gemma}. The plan-validity reward model is a separate smaller generative verifier (Qwen3.5-0.8B, \S\ref{sec:llm-judge-rm}) frozen during warm-up and CPPO. We choose its decision threshold $\tau$ in \eqref{eq:jpsi} on a validation set to obtain a high-recall validity gate; the main experiments use $\tau{=}0.17$ and $M{=}8$ planner tuples per prompt for the across-tuple advantage of \eqref{eq:planner-adv}. Trainable parameters use full precision: at CPPO-scale learning rates the AdamW~\citep{loshchilov2019adamw} update falls below the bf16 mantissa resolution, rendering half-precision optimization numerically ineffective. The frozen reference and reward models stay in half precision.

\subsection{Training Data}
\label{sec:train-data}

All training draws from CodeContests~\citep{li2022competition}:
planner SFT, reward-model data, the planner warm-up, and the joint
CPPO rollout pool all use CodeContests train, whose problems admit
enough alternative solution strategies to support tuple-level
coordination. APPS~\citep{hendrycks2021apps} and LiveCodeBench-v6
(LCBv6)~\citep{jain2024livecodebench} are \emph{evaluation-only}:
neither ever enters any training-side pool---no planner SFT, no
LLM-judge label, no reward-model train/val, no warm-up, no CPPO
rollout, no hyperparameter sweep, no qualitative example---so both are
held-out cross-corpus transfer benchmarks, while CodeContests valid
measures in-domain pass@$K$. Table~\ref{tab:data-splits} lists the source split used by
each stage; problem counts and the full per-component exposure audit
are in Appendix~\ref{app:data-splits}
(Table~\ref{tab:data-splits-counts}).

\begin{table}[t]
\centering
\caption{Stage-wise data sources. Held-out evaluation sets are
disjoint from all training-side pools under the decontamination
protocol in Appendix~\ref{app:data-splits}.}
\scriptsize
\setlength{\tabcolsep}{4pt}
\begin{tabular}{@{}p{0.34\linewidth}p{0.18\linewidth}p{0.40\linewidth}@{}}
\toprule
Stage & Role & Source split \\
\midrule
Planner SFT & train & CodeContests train \\
Reward model & train/val & CodeContests train, filtered \\
Planner warm-up & train & CodeContests train, filtered \\
Joint CPPO & train & CodeContests train \\
\midrule
APPS & eval & APPS test \\
CodeContests & eval & CodeContests valid \\
LCBv6 & eval & LCBv6 held-out \\
\bottomrule
\end{tabular}
\label{tab:data-splits}
\end{table}

\paragraph{Decontamination protocol.}
We deduplicate each corpus before splitting. If a training-side item matches a
held-out evaluation problem, we remove it from training and keep the evaluation
split fixed. We match problems by official identifiers when available
(Codeforces contest/index, APPS id, LCBv6 id), and otherwise by a canonicalized
statement hash with fuzzy $n$-gram matching (Appendix~\ref{app:decon}). After
filtering, all APPS--CodeContests, APPS--LCBv6, and CodeContests--LCBv6
train--evaluation pairs have zero prompt-hash overlap.

\paragraph{Candidate selection funnel.}
Planner SFT uses an over-generate--then-filter pipeline over self-generated plan samples; the
candidate-generation prompt is in Appendix~\ref{app:prompts} and the
rest of the construction is in Appendix~\ref{app:sft-funnel}.

\subsection{Evaluation}
\label{sec:evaluation}
\label{sec:setup-eval}

\paragraph{Metric and benchmarks.}
Our primary metric is pass@$K$ under a fixed verifier budget on three
competitive-programming benchmarks: APPS~\citep{hendrycks2021apps},
CodeContests~\citep{li2022competition}, and LiveCodeBench-v6
(LCBv6)~\citep{jain2024livecodebench}. A CPPO rollout consumes one
planner call and $K$ solver calls, so we report both
verifier/solver-attempt pass@$K$ (matching prior work) and
token-normalized pass@$K$ (Table~\ref{tab:decoded-token-efficiency}),
which accounts for the planner-call overhead. The full APPS pass@$K$
sweep across baselines and model sizes is in
Appendix~\ref{app:passk-sweep}.

\paragraph{Verifier protocol.}
Candidate programs are extracted with the same parser for every method
and executed in an isolated sandbox. Timeout, memory overflow,
compilation error, runtime error, empty extraction, and wrong answer
are all scored as $V(x, y){=}0$. We use the official benchmark test
cases when available, and never expose evaluation tests during
training. Sandbox configuration (timeout, memory cap, network and
filesystem isolation) is reported in Appendix~\ref{app:sandbox}.

\paragraph{Baselines.}
CPPO is compared against (i) independent answer sampling at the same $K$;
(ii) independent sketch-then-solve, where each sketch is sampled without
seeing the others; (iii) self-consistency and other inference-time
aggregation \citep{wang2022selfconsistency}; (iv) planning and search
baselines---Plan-and-Solve \citep{wang2023plan} and PlanSearch
\citep{wang2024plansearch}; (v) pass@$K$ training without a planner reward tied to tuple
validity and solver success \citep{chen2025passk,walder2025pkpo}; and
(vi) ablations that remove the
plan-validity reward or the outcome reward from the planner.

\paragraph{Token accounting.}
Table~\ref{tab:decoded-token-efficiency} reports token-normalized
pass@$4$ on the held-out APPS and LiveCodeBench-v6 evaluation sets
(Table~\ref{tab:data-splits-counts}). On these sets, the
concise \textsc{plan} contract yields shorter
strategy-conditioned solver completions, more than offsetting the planner
generation in decoded-token accounting (planner--solver breakdown in
Appendix~\ref{app:decoded-token-accounting}).

\begin{table}[t]
\centering
\caption{Decoded-token-normalized pass@$4$ for Qwen3.5-4B at $K{=}4$
on the held-out APPS and LiveCodeBench-v6 evaluation sets (failed and
empty branches included). Prompt tokens are excluded; full accounting in
Appendix~\ref{app:decoded-token-accounting}.}
\label{tab:decoded-token-efficiency}
\scriptsize
\setlength{\tabcolsep}{3pt}
\renewcommand{\arraystretch}{0.95}
\begin{tabular}{llccc}
\toprule
Dataset & Method & pass@4 & Decoded toks. & pass@4 / 10k toks. \\
\midrule
APPS & Direct Solve      & 0.515 & 6087 & 0.846 \\
APPS & Plan-and-Solve    & 0.530 & 7326 & 0.723 \\
APPS & PlanSearch  & 0.554 & 5535 & 1.001 \\
APPS & \textbf{CPPO}     & \textbf{0.784} & \textbf{4728} & \textbf{1.658} \\
\midrule
LCBv6 & Direct Solve      & 0.214 & 1731 & 1.236 \\
LCBv6 & Plan-and-Solve    & 0.255 & 3975 & 0.642 \\
LCBv6 & PlanSearch  & 0.236 & 2137 & 1.104 \\
LCBv6 & PKPO        & 0.488 & 2137 & 2.284 \\
LCBv6 & UpSkill     & 0.411 & 2218 & 1.853 \\
LCBv6 & \textbf{CPPO}     & \textbf{0.505} & \textbf{1556} & \textbf{3.246} \\
\bottomrule
\end{tabular}
\end{table}

\paragraph{Variance and significance.}
For training-based methods---Tuple Planner SFT, RM warm-up, Full
CPPO, Pass@$K$ Training / RLVR, PKPO, and UpSkill---we report
$\mathrm{mean}_{\pm\,\mathrm{std}}$ over three independent training
seeds; each seed re-runs the relevant parameter-updating stages
end-to-end under the same data splits and hyperparameters. For
inference-only baselines that we run ourselves (Direct Solve,
Plan-and-Solve, PlanSearch), variation is over evaluation problems
only. Pairwise significance ($\dagger$) uses a \emph{hierarchical
bootstrap} that resamples both evaluation problems and training seeds:
each of $1000$ resamples draws problems with replacement and, for each
trained method, draws one of its three seeds with replacement, so the
reported $p$-values reflect evaluation noise \emph{and} training-seed
variability rather than being conditional on a fixed checkpoint. With
only three seeds this seed-level test is deliberately conservative;
Appendix~\ref{app:seed-significance} additionally reports the per-seed
CPPO-minus-strongest-baseline difference, its mean, and a $95\%$
confidence interval over seeds. Externally reported numbers, such as
the LCBv6 Direct Solve cells of Gemma~4 in Table~\ref{tab:gemma}, use
the uncertainty provided by the original source when available;
otherwise we report the point estimate and exclude those cells from
significance tests.

\subsection{Validity gate diagnostics}
\begin{figure}[!th]
  \vspace{-0.5ex}
\centering
\begin{subfigure}[b]{0.48\linewidth}
\centering
\includegraphics[width=\linewidth]{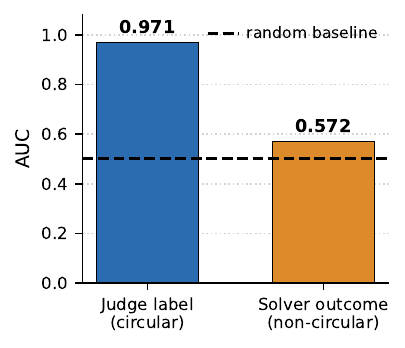}
\caption{RM agreement}
\label{fig:rm-panel-a}
\end{subfigure}\hfill
\vspace{-0.5ex}
\begin{subfigure}[b]{0.48\linewidth}
\centering
\includegraphics[width=\linewidth]{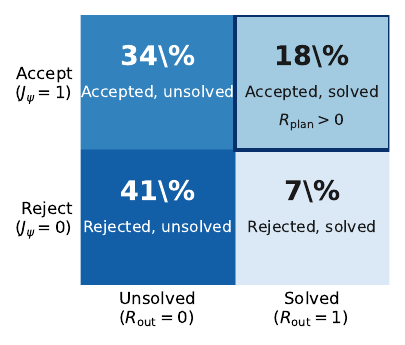}
\caption{Reward decomposition}
\label{fig:rm-panel-b}
\end{subfigure}
\caption{Reward-model diagnostics.
(a) The RM matches held-out judge labels but weakly predicts
frozen-solver outcomes, supporting its use as a validity gate rather
than a plan-quality scorer.
(b) Joint distribution of validity decisions and solver outcomes during
CPPO rollouts; shading encodes rollout frequency, and only the
\emph{accepted, solved} cell yields nonzero $\Rplan$.}
\label{fig:rm-outcome-gap}
\vspace{-0.5ex}
\end{figure}

\vspace{-0.4ex}
\label{sec:validity-gate-diag}
\label{sec:rm-gate-rationale}
The plan-validity RM fits held-out LLM-judge labels well
(AUC $0.971$; Figure~\ref{fig:rm-outcome-gap}a) but only weakly predicts
frozen-solver pass@$K$ outcomes (AUC $0.572$). At $\tau{=}0.17$, however,
accepted tuples pass more often than rejected ones ($35\%$ vs. $15\%$), and
only accepted-and-solved tuples receive nonzero $\Rplan$
(Figure~\ref{fig:rm-outcome-gap}b). Thus $\Jpsi$ serves as a validity
filter: it suppresses malformed or duplicated tuples and plans that leak the
answer, as shown by the duplicate-rate jump under $-\Jpsi$ in
Table~\ref{tab:ablation-components} ($0.08{\to}0.32$). It is too noisy to rank
surviving plans by solver utility, so we use it only to gate training rollouts
and leave $\Rout$ as the quality signal.

\subsection{Ablation}
\label{sec:ablation}

Table~\ref{tab:ablation} isolates the contribution of each stage of
the CPPO pipeline. Each row adds one stage on top of the previous one:
Direct Solve, planner-only SFT, planner SFT followed by RM-guided
warm-up, and the full CPPO run.

\begin{table*}[!h]
\centering
\caption{Ablation over the CPPO pipeline on APPS~\citep{hendrycks2021apps},
CodeContests~\citep{li2022competition}, and LiveCodeBench-v6
(LCBv6)~\citep{jain2024livecodebench}. All numbers are pass@$4$.}
\label{tab:ablation}
\scriptsize
\setlength{\tabcolsep}{5pt}
\begin{tabular}{@{}llccc@{}}
\toprule
Method & Base model & APPS & CodeContests & LiveCodeBench~(LCBv6) \\
\midrule
Direct Solve & Qwen3.5-2B & 0.420\std{0.017} & 0.040\std{0.012} & 0.161\std{0.014} \\
Direct Solve & Qwen3.5-4B & 0.515\std{0.022} & 0.094\std{0.014} & 0.214\std{0.013} \\
Direct Solve & Qwen3.5-9B & 0.696\std{0.021} & 0.175\std{0.019} & 0.481\std{0.014} \\
\midrule
Tuple Planner SFT & Qwen3.5-2B & 0.453\std{0.020} & 0.087\std{0.020} & 0.232\std{0.017} \\
Tuple Planner SFT & Qwen3.5-4B & 0.548\std{0.024} & 0.147\std{0.021} & 0.348\std{0.018} \\
Tuple Planner SFT & Qwen3.5-9B & 0.733\std{0.016} & 0.279\std{0.026} & 0.514\std{0.015} \\
\midrule
Tuple Planner SFT + RM warm-up & Qwen3.5-2B & 0.461\std{0.024} & 0.105\std{0.018} & 0.231\std{0.015} \\
Tuple Planner SFT + RM warm-up & Qwen3.5-4B & 0.564\std{0.023} & 0.204\std{0.021} & 0.384\std{0.013} \\
Tuple Planner SFT + RM warm-up & Qwen3.5-9B & 0.744\std{0.019} & 0.313\std{0.021} & 0.541\std{0.018} \\
\midrule
\textbf{CPPO} & Qwen3.5-2B & 0.536\std{0.017} & 0.201\std{0.021} & 0.418\std{0.017} \\
\textbf{CPPO} & Qwen3.5-4B & 0.784\std{0.013} & 0.231\std{0.025} & 0.505\std{0.017} \\
\textbf{CPPO} & Qwen3.5-9B & 0.806\std{0.019} & 0.396\std{0.030} & 0.728\std{0.016} \\
\bottomrule
\end{tabular}
\end{table*}

The ablation separates three effects. First, planner-only SFT is not a
complete explanation for CPPO's gains: it improves over Direct Solve on
every cell, but by a margin far smaller than full CPPO.
Second, RM-guided warm-up is most useful on CodeContests and is modest or
mixed elsewhere, consistent with its role as a validity constraint rather
than a task-quality signal. Third, the full CPPO update improves over the
warm-up stage in every model--dataset cell, indicating that the main gains
come from joint planner--solver RL with verifier-confirmed outcome credit.

\subsubsection{Reward-component and rollout ablations}
\label{sec:component-ablation}

Table~\ref{tab:ablation-components} isolates CPPO's two main design choices:
the validity gate and across-tuple planner normalization. Removing $\Jpsi$
quadruples the duplicate-plan rate while leaving pass@$4$ close to full CPPO,
showing that the gate mainly suppresses invalid planner credit rather than
scoring task quality. The $\Jpsi$-only variant removes outcome feedback, while
$M{=}1$ removes across-tuple contrast by replacing \eqref{eq:planner-adv} with a
single-sample REINFORCE update. Increasing $M$ improves the across-tuple
advantage with diminishing returns by $M{=}8$ (we did not test beyond). The bottom block sweeps
$K\in\{2,4,8\}$ using the same $\Ktuple{=}4$ planner from
Table~\ref{tab:main}, with truncation for $K{<}4$ and pooling for $K{>}4$.
Appendix~\ref{app:passk-sweep} gives the full larger-budget decomposition, and
Appendix~\ref{app:inference-mode} reports the matched joint-vs.-iid
plan-sampling ablation.

\begin{table}[thb]
\centering
\caption{Planner-reward ablations and pass@$K$ sweep on APPS /
Qwen3.5-2B. The top block contains two sub-experiments: the
$-\Jpsi$, $\Jpsi$-only rows vary the \emph{reward} (relative to Full
CPPO, $\Rplan = \Jpsi \cdot \Rout$, $M{=}8$), while the
$M{\in}\{1,2,4\}$ rows hold the full $\Rplan = \Jpsi \cdot \Rout$
reward fixed and vary only the across-tuple sample count $M$. The
bottom block varies the solver-attempt budget $K$. Duplicate rate
measures duplicate-plan tuples; lower is better. Uncertainties follow
the convention of \S\ref{sec:evaluation}.}
\label{tab:ablation-components}
\scriptsize
\setlength{\tabcolsep}{3pt}
\renewcommand{\arraystretch}{0.78}
\begin{tabular}{@{}lcc@{}}
\toprule
\multicolumn{3}{@{}l}{\textit{Planner-reward variants ($M{=}8$)}} \\
\midrule
Variant & pass@4 $\uparrow$ & Dup. rate $\downarrow$ \\
\midrule
Full CPPO ($J_\psi R_{\mathrm{out}}$)
  & $\mathbf{0.536}\pm0.017$ & $\mathbf{0.08}\pm0.030$ \\
$-J_\psi$ gate ($R_{\mathrm{out}}$)
  & $0.522\pm0.023$ & $0.32\pm0.044$ \\
$J_\psi$-only
  & $0.503\pm0.024$ & $0.14\pm0.032$ \\
\midrule
\multicolumn{3}{@{}l}{\textit{Across-tuple sample count $M$ ($R_{\mathrm{plan}} = J_\psi \cdot R_{\mathrm{out}}$)}} \\
\midrule
$M{=}4$
  & $0.535\pm0.023$ & $0.10\pm0.032$ \\
$M{=}2$
  & $0.515\pm0.022$ & $0.12\pm0.034$ \\
$M{=}1$
  & $0.486\pm0.022$ & $0.16\pm0.035$ \\
\midrule
\multicolumn{3}{@{}l}{\textit{Attempt-budget sweep}} \\
\midrule
$K$ & Direct Solve & CPPO \\
\midrule
$2$ & $0.399\pm0.022$ & $\mathbf{0.518}\pm0.021$ \\
$4$ & $0.420\pm0.017$ & $\mathbf{0.536}\pm0.017$ \\
$8$ & $0.432\pm0.024$ & $\mathbf{0.561}\pm0.024$ \\
\bottomrule
\end{tabular}
\end{table}

\begin{table*}[tbh]
\centering
\caption{Majority-vote accuracy~(maj@$K$) at $K{=}4$ on the
LiveCodeBench output-prediction tasks---code execution and test-output
prediction---for Qwen3.5-2B and Qwen3.5-4B. For each problem we draw
four samples and score the majority answer against the canonical
output.}
\label{tab:majvote}
\scriptsize
\setlength{\tabcolsep}{5pt}
\begin{tabular}{@{}llccc@{}}
\toprule
Method & Base model & Code Execution & Test Output & Avg. \\
\midrule
Direct                        & Qwen3.5-2B & 0.22\std{0.041} & 0.19\std{0.042} & 0.205\std{0.042} \\
Plan-and-Solve                & Qwen3.5-2B & 0.16\std{0.039} & 0.10\std{0.033} & 0.130\std{0.031} \\
Fixed-Strategy Ensemble       & Qwen3.5-2B & 0.19\std{0.042} & 0.13\std{0.034} & 0.160\std{0.033} \\
PKPO                          & Qwen3.5-2B & 0.38\std{0.050} & 0.21\std{0.044} & 0.295\std{0.042} \\
\textbf{CPPO}                 & Qwen3.5-2B & \textbf{0.41}\std{0.049} & \textbf{0.31}\std{0.047} & \textbf{0.360}\std{0.046} \\
\midrule
Direct                        & Qwen3.5-4B & 0.38\std{0.052} & 0.29\std{0.044} & 0.335\std{0.045} \\
Plan-and-Solve                & Qwen3.5-4B & 0.15\std{0.034} & 0.12\std{0.030} & 0.135\std{0.035} \\
Fixed-Strategy Ensemble       & Qwen3.5-4B & 0.32\std{0.048} & 0.30\std{0.049} & 0.310\std{0.042} \\
PKPO                          & Qwen3.5-4B & 0.48\std{0.048} & 0.34\std{0.048} & 0.410\std{0.052} \\
\textbf{CPPO}                 & Qwen3.5-4B & \textbf{0.51}\std{0.054} & \textbf{0.35}\std{0.050} & \textbf{0.430}\std{0.052} \\
\bottomrule
\end{tabular}
\end{table*}

\subsection{Majority-vote consistency~(maj@$4$)}
\label{sec:majvote}

Pass@$K$ rewards \emph{any} successful branch, so a method that spreads
probability mass across strategies could in principle inflate pass@$K$
while leaving the modal answer unchanged---or worse, diluting it on
tasks with a single canonical output. We test this directly on the
LiveCodeBench output-prediction subsets (code-execution and test-output
prediction), where each problem has exactly one correct answer and
maj@$K$ is well-defined: for each problem we draw $K{=}4$ samples and
score the most frequent output against the canonical label, breaking
ties by sample order. Table~\ref{tab:majvote} reports the results for
Qwen3.5-2B and Qwen3.5-4B.

CPPO wins both sub-tasks at both sizes: on Qwen3.5-2B, average
maj@$4$ of $0.360$ versus $0.205$ for Direct Solve and $0.295$ for
PKPO; on Qwen3.5-4B, $0.430$ versus $0.335$ and $0.410$ respectively.
Coordinated planning tightens rather than diffuses the modal answer:
diversity across high-level strategies is consistent with sharper
outputs once a strategy is chosen.

\paragraph{Generality across base models.}
\label{sec:generality-gemma}
The same trend holds on Gemma~4 E2B and E4B~\citep{google2026gemma4}
under an identical pipeline (Appendix~\ref{app:gemma}).

\section{Conclusion}
\label{sec:conclusion}
We presented CPPO, a planner--solver method that allocates a fixed pass@$K$ budget to complementary solution strategies rather than repeated samples from one answer distribution. By tying planner credit to both plan validity and verifier-confirmed solver success, CPPO encourages its $K$ attempts to cover different algorithmic strategies rather than paraphrases of a single approach. Across APPS, CodeContests, and LiveCodeBench-v6, it improves pass@$4$ over direct sampling, planning baselines, planner-only SFT, and pass@$K$-oriented RL on Qwen3.5-\{2B, 4B, 9B\}, with the same trend carrying over to Gemma~4 E2B and E4B. Future improvements in pass@$K$ may depend less on sampling more answers and more on coordinating which strategies those answers pursue.

% EMNLP/ACL: Limitations is a required, unnumbered section placed after the
% conclusion and before the references. It does not count toward the page limit.
\vspace{-1ex}
\section*{Limitations}
\vspace{-0.4ex}
\paragraph{Scope.}
Our claims are restricted to competitive-programming benchmarks, where
problems admit several distinct algorithmic strategies. Math benchmarks
such as MATH or AIME usually have a single canonical solution path and
are outside our scope---there, CPPO may reduce to a more expensive
form of plan-and-solve.

\vspace{-0.4ex}
\paragraph{Dependencies.}
The planner reward $\Rplan = \Jpsi \cdot \Rout$ is sparse by construction, so CPPO requires the warm-up and audit gate to train at all. It also inherits the biases of the offline LLM judge and uses a shared planner--solver backbone; splitting them is left to future work.

% Ethics Statement: optional, unnumbered, also page-limit-exempt.
\vspace{-0.4ex}
\section*{Ethics Statement}
\vspace{-0.4ex}
CPPO trains code policies with execution-based rewards. The verifier checks functional correctness only, not security or robustness, so downstream use requires review. All generated code runs in a sandboxed, time-bounded runner. We use only public APPS, CodeContests, and LiveCodeBench-v6 data under their original licenses and no PII. The plan-validity reward model may inherit biases from the offline LLM judge, so we validate it on balanced held-out data before use.

\vspace{-0.4ex}
\paragraph{Code and artifact release.}
\vspace{-0.4ex}
Upon acceptance, we will release code, RM checkpoints, judge prompts, and sandbox configuration, but not benchmark content.

% acl.sty already sets \bibliographystyle{acl_natbib}.
% Wrap the bibliography in \sloppy so long unbreakable author runs in the
% auto-generated .bbl do not raise Underfull \hbox warnings.
{\sloppy
\bibliography{bibliography}

@article{chen2021codex,
  title={Evaluating large language models trained on code},
  author={Chen, Mark and Tworek, Jerry and Jun, Heewoo and Yuan, Qiming and Pinto, Henrique Ponde de Oliveira and Kaplan, Jared and Edwards, Harri and Burda, Yuri and Joseph, Nicholas and Brockman, Greg and others},
  journal={arXiv preprint arXiv:2107.03374},
  year={2021}
}

@inproceedings{wang2022selfconsistency,
  title={Self-consistency improves chain of thought reasoning in language models},
  author={Wang, Xuezhi and Wei, Jason and Schuurmans, Dale and Le, Quoc and Chi, Ed and Narang, Sharan and Chowdhery, Aakanksha and Zhou, Denny},
  booktitle={International Conference on Learning Representations (ICLR)},
  year={2023}
}

@inproceedings{wang2023plan,
  title={Plan-and-solve prompting: Improving zero-shot chain-of-thought reasoning by large language models},
  author={Wang, Lei and Xu, Wanyu and Lan, Yihuai and Hu, Zhiqiang and Lan, Yunshi and Lee, Roy Ka-Wei and Lim, Ee-Peng},
  booktitle={Proceedings of the 61st Annual Meeting of the Association for Computational Linguistics (Volume 1: Long Papers)},
  pages={2609--2634},
  year={2023}
}

@inproceedings{yao2023tot,
  title={Tree of thoughts: Deliberate problem solving with large language models},
  author={Yao, Shunyu and Yu, Dian and Zhao, Jeffrey and Shafran, Izhak and Griffiths, Thomas L. and Cao, Yuan and Narasimhan, Karthik},
  booktitle={Advances in Neural Information Processing Systems (NeurIPS)},
  year={2023}
}

@article{shao2024deepseekmath,
  title={{DeepSeekMath}: Pushing the limits of mathematical reasoning in open language models},
  author={Shao, Zhihong and Wang, Peiyi and Zhu, Qihao and Xu, Runxin and Song, Junxiao and Bi, Xiao and Zhang, Haowei and Zhang, Mingchuan and Li, Y. K. and Wu, Y. and Guo, Daya},
  journal={arXiv preprint arXiv:2402.03300},
  year={2024}
}

@article{wang2024plansearch,
  title={Planning in natural language improves {LLM} search for code generation},
  author={Wang, Evan and Cassano, Federico and Wu, Catherine and Bai, Yunfeng and Song, Will and Nath, Vaskar and Han, Ziwen and Hendryx, Sean and Yue, Summer and Zhang, Hugh},
  journal={arXiv preprint arXiv:2409.03733},
  year={2024}
}

@article{chen2025passk,
  title={Pass@k training for adaptively balancing exploration and exploitation of large reasoning models},
  author={Chen, Zhipeng and Qin, Xiaobo and Wu, Youbin and Ling, Yue and Ye, Qinghao and Zhao, Wayne Xin and Shi, Guang},
  journal={arXiv preprint arXiv:2508.10751},
  year={2025}
}

@inproceedings{hendrycks2021apps,
  title={Measuring Coding Challenge Competence with {APPS}},
  author={Hendrycks, Dan and Basart, Steven and Kadavath, Saurav and Mazeika, Mantas and Arora, Akul and Guo, Ethan and Burns, Collin and Puranik, Samir and He, Horace and Song, Dawn and Steinhardt, Jacob},
  booktitle={Advances in Neural Information Processing Systems (NeurIPS) Datasets and Benchmarks Track},
  year={2021}
}

@article{li2022competition,
  title={Competition-level code generation with {AlphaCode}},
  author={Li, Yujia and Choi, David and Chung, Junyoung and Kushman, Nate and Schrittwieser, Julian and Leblond, Remi and Eccles, Tom and Keeling, James and Gimeno, Felix and Dal Lago, Agustin and Hubert, Thomas and Choy, Peter and de Masson d'Autume, Cyprien and Babuschkin, Igor and Chen, Xinyun and Huang, Po-Sen and Welbl, Johannes and Gowal, Sven and Cherepanov, Alexey and Molloy, James and Mankowitz, Daniel J. and Sutherland Robson, Esme and Kohli, Pushmeet and de Freitas, Nando and Kavukcuoglu, Koray and Vinyals, Oriol},
  journal={Science},
  volume={378},
  number={6624},
  pages={1092--1097},
  year={2022}
}

@article{walder2025pkpo,
  title={Pass@k Policy Optimization: Solving Harder Reinforcement Learning Problems},
  author={Walder, Christian and Karkhanis, Deep},
  journal={arXiv preprint arXiv:2505.15201},
  year={2025}
}

@article{shah2026upskill,
  title={{UpSkill}: Mutual Information Skill Learning for Structured Response Diversity in {LLM}s},
  author={Shah, Devan and Yang, Owen and Yang, Daniel and Zheng, Chongyi and Eysenbach, Benjamin},
  journal={arXiv preprint arXiv:2602.22296},
  year={2026}
}

@inproceedings{jain2024livecodebench,
  title={{LiveCodeBench}: Holistic and Contamination Free Evaluation of Large Language Models for Code},
  author={Jain, Naman and Han, King and Gu, Alex and Li, Wen-Ding and Yan, Fanjia and Zhang, Tianjun and Wang, Sida and Solar-Lezama, Armando and Sen, Koushik and Stoica, Ion},
  booktitle={International Conference on Learning Representations (ICLR)},
  year={2025}
}

@article{brown2024monkeys,
  title={Large Language Monkeys: Scaling Inference Compute with Repeated Sampling},
  author={Brown, Bradley and Juravsky, Jordan and Ehrlich, Ryan and Clark, Ronald and Le, Quoc V. and R{\'e}, Christopher and Mirhoseini, Azalia},
  journal={arXiv preprint arXiv:2407.21787},
  year={2024}
}

@inproceedings{snell2024scaling,
  title={Scaling {LLM} Test-Time Compute Optimally can be More Effective than Scaling Model Parameters},
  author={Snell, Charlie and Lee, Jaehoon and Xu, Kelvin and Kumar, Aviral},
  booktitle={International Conference on Learning Representations (ICLR)},
  year={2025}
}

@article{guo2025deepseekr1,
  title={{DeepSeek-R1}: Incentivizing Reasoning Capability in {LLMs} via Reinforcement Learning},
  author={Guo, Daya and Yang, Dejian and Zhang, Haowei and others},
  journal={Nature},
  volume={645},
  pages={633--638},
  doi={10.1038/s41586-025-09422-z},
  year={2025}
}

@article{li2025darling,
  title={Jointly Reinforcing Diversity and Quality in Language Model Generations},
  author={Li, Tianjian and Zhang, Yiming and Yu, Ping and Saha, Swarnadeep and Khashabi, Daniel and Weston, Jason and Lanchantin, Jack and Wang, Tianlu},
  journal={arXiv preprint arXiv:2509.02534},
  year={2025}
}

@inproceedings{hu2025diver,
  title={Diversity-Incentivized Exploration for Versatile Reasoning},
  author={Hu, Zican and Zhang, Shilin and Li, Yafu and Yan, Jianhao and Hu, Xuyang and Cui, Leyang and Qu, Xiaoye and Chen, Chunlin and Cheng, Yu and Wang, Zhi},
  booktitle={International Conference on Learning Representations (ICLR)},
  year={2026}
}

@inproceedings{lightman2023prm,
  title={Let's Verify Step by Step},
  author={Lightman, Hunter and Kosaraju, Vineet and Burda, Yura and Edwards, Harri and Baker, Bowen and Lee, Teddy and Leike, Jan and Schulman, John and Sutskever, Ilya and Cobbe, Karl},
  booktitle={International Conference on Learning Representations (ICLR)},
  year={2024}
}

@article{cobbe2021verifiers,
  title={Training Verifiers to Solve Math Word Problems},
  author={Cobbe, Karl and Kosaraju, Vineet and Bavarian, Mohammad and Chen, Mark and Jun, Heewoo and Kaiser, Lukasz and Plappert, Matthias and Tworek, Jerry and Hilton, Jacob and Nakano, Reiichiro and Hesse, Christopher and Schulman, John},
  journal={arXiv preprint arXiv:2110.14168},
  year={2021}
}

@inproceedings{zheng2023judge,
  title={Judging {LLM-as-a-Judge} with {MT-Bench} and Chatbot Arena},
  author={Zheng, Lianmin and Chiang, Wei-Lin and Sheng, Ying and Zhuang, Siyuan and Wu, Zhanghao and Zhuang, Yonghao and Lin, Zi and Li, Zhuohan and Li, Dacheng and Xing, Eric P. and Zhang, Hao and Gonzalez, Joseph E. and Stoica, Ion},
  booktitle={Advances in Neural Information Processing Systems (NeurIPS)},
  year={2023}
}

@article{schulman2017ppo,
  title={Proximal Policy Optimization Algorithms},
  author={Schulman, John and Wolski, Filip and Dhariwal, Prafulla and Radford, Alec and Klimov, Oleg},
  journal={arXiv preprint arXiv:1707.06347},
  year={2017}
}

@inproceedings{loshchilov2019adamw,
  title={Decoupled Weight Decay Regularization},
  author={Loshchilov, Ilya and Hutter, Frank},
  booktitle={International Conference on Learning Representations (ICLR)},
  year={2019}
}

@misc{qwen2026qwen35,
  title={{Qwen3.5} Open Models},
  author={{Qwen Team}},
  year={2026},
  howpublished={Hugging Face model collection, \url{https://huggingface.co/Qwen}},
  note={Qwen3.5 series: 2B, 4B, and 9B base and chat variants. Exact checkpoint identifiers used in this paper are listed in Appendix~\ref{app:models-checkpoints}. Accessed 2026-05-26.}
}

@misc{google2026gemma4,
  title={{Gemma~4}: Open Lightweight Models},
  author={{Google DeepMind}},
  year={2026},
  howpublished={Official model cards, \url{https://ai.google.dev/gemma}},
  note={Gemma~4 E2B and E4B. LiveCodeBench-v6 pass@$4$ values used in this paper ($44.0\%$ for E2B, $52.0\%$ for E4B) reproduced from the official model-card evaluation tables. Accessed 2026-05-26.}
}

@inproceedings{zhu2018texygen,
  title={{Texygen}: A Benchmarking Platform for Text Generation Models},
  author={Zhu, Yaoming and Lu, Sidi and Zheng, Lei and Guo, Jiaxian and Zhang, Weinan and Wang, Jun and Yu, Yong},
  booktitle={Proceedings of the 41st International ACM SIGIR Conference on Research \& Development in Information Retrieval},
  pages={1097--1100},
  year={2018}
}
\par}

\clearpage
\appendix
\section{Main Results: Full Table Across All Base-Model Sizes}
\label{app:main-full}

Table~\ref{tab:main-full} reproduces Table~\ref{tab:main} of the main
text with the Qwen3.5-2B rows included. The 2B numbers track the
4B/9B trends already discussed in \S\ref{sec:intro} but with smaller
absolute deltas; the three unmarked cells---2B APPS, 4B LCBv6, and 9B
APPS---are where the CPPO advantage stays positive but does not pass
the hierarchical problem-and-seed bootstrap threshold.

\emph{Tuple Planner SFT} samples the $K$ plans jointly from
$q_\Theta(s_i \mid x, s_{<i})$ (the joint-tuple policy class of
\S\ref{sec:design}, evaluated without RL); the matching iid-inference
variant on the same checkpoint is reported in
Appendix~\ref{app:inference-mode} as an ablation on autoregressive
conditioning.

\begin{table*}[!h]
\centering
\caption{Pass@$4$ results under the same $K{=}4$ solver-attempt budget and
verifier across base model sizes. Full version of
Table~\ref{tab:main} of the main text, with the Qwen3.5-2B rows
included. \emph{Tuple Planner SFT} samples the $K$ plans jointly from
$q_\Theta(s_i \mid x, s_{<i})$; CPPO trains the joint planner--solver
policy on this tuple class and gates planner credit on a learned
plan-validity reward.}
\label{tab:main-full}
\scriptsize
\setlength{\tabcolsep}{3pt}
\renewcommand{\arraystretch}{0.8}
\begin{tabular}{@{}llccc@{}}
\toprule
Method & Base model & APPS~\citep{hendrycks2021apps} & \shortstack{CodeContests~\citep{li2022competition}} & \shortstack{LiveCodeBench~(LCBv6)~\citep{jain2024livecodebench}} \\
\midrule
Direct Solve & Qwen3.5-2B & 0.420\std{0.017} & 0.040\std{0.012} & 0.161\std{0.014} \\
Direct Solve & Qwen3.5-4B & 0.515\std{0.022} & 0.094\std{0.014} & 0.214\std{0.013} \\
Direct Solve & Qwen3.5-9B & 0.696\std{0.021} & 0.175\std{0.019} & 0.481\std{0.014} \\
Plan-and-Solve & Qwen3.5-2B & 0.492\std{0.021} & 0.065\std{0.012} & 0.205\std{0.014} \\
Plan-and-Solve & Qwen3.5-4B & 0.530\std{0.019} & 0.098\std{0.016} & 0.255\std{0.012} \\
Plan-and-Solve & Qwen3.5-9B & 0.801\std{0.015} & 0.171\std{0.017} & 0.490\std{0.020} \\
PlanSearch & Qwen3.5-2B & 0.513\std{0.016} & 0.043\std{0.012} & 0.225\std{0.015} \\
PlanSearch & Qwen3.5-4B & 0.554\std{0.022} & 0.128\std{0.021} & 0.236\std{0.012} \\
PlanSearch & Qwen3.5-9B & 0.770\std{0.019} & 0.168\std{0.018} & 0.564\std{0.019} \\
Pass@$K$ Training / RLVR & Qwen3.5-2B & 0.476\std{0.021} & 0.085\std{0.015} & 0.215\std{0.015} \\
Pass@$K$ Training / RLVR & Qwen3.5-4B & 0.607\std{0.017} & 0.115\std{0.020} & 0.316\std{0.019} \\
Pass@$K$ Training / RLVR & Qwen3.5-9B & 0.762\std{0.015} & 0.185\std{0.021} & 0.503\std{0.020} \\
PKPO & Qwen3.5-2B & 0.448\std{0.024} & 0.062\std{0.012} & 0.347\std{0.016} \\
PKPO & Qwen3.5-4B & 0.722\std{0.021} & 0.156\std{0.020} & 0.488\std{0.015} \\
PKPO & Qwen3.5-9B & 0.787\std{0.020} & 0.183\std{0.025} & 0.588\std{0.015} \\
UpSkill & Qwen3.5-2B & 0.462\std{0.018} & 0.041\std{0.012} & 0.287\std{0.018} \\
UpSkill & Qwen3.5-4B & 0.661\std{0.023} & 0.121\std{0.019} & 0.411\std{0.016} \\
UpSkill & Qwen3.5-9B & 0.762\std{0.015} & 0.136\std{0.017} & 0.542\std{0.013} \\
\midrule
Tuple Planner SFT & Qwen3.5-2B & 0.453\std{0.020} & 0.087\std{0.020} & 0.232\std{0.017} \\
Tuple Planner SFT & Qwen3.5-4B & 0.548\std{0.024} & 0.147\std{0.021} & 0.348\std{0.018} \\
Tuple Planner SFT & Qwen3.5-9B & 0.733\std{0.016} & 0.279\std{0.026} & 0.514\std{0.015} \\
\textbf{CPPO} & Qwen3.5-2B & 0.536\std{0.017} & 0.201$^{\dagger}$\std{0.021} & 0.418$^{\dagger}$\std{0.017} \\
\textbf{CPPO} & Qwen3.5-4B & 0.784$^{\dagger}$\std{0.013} & 0.231$^{\dagger}$\std{0.025} & 0.505\std{0.017} \\
\textbf{CPPO} & Qwen3.5-9B & 0.806\std{0.019} & 0.396$^{\dagger}$\std{0.030} & 0.728$^{\dagger}$\std{0.016} \\
\bottomrule
\end{tabular}

\smallskip
\begin{minipage}{\textwidth}
\scriptsize
$^{\dagger}$ CPPO significantly outperforms the strongest non-CPPO
baseline in that cell at $p < 0.05$ under a hierarchical bootstrap over
problems and training seeds ($1000$ resamples; per-seed differences and
$95\%$ CIs in Appendix~\ref{app:seed-significance}). Unmarked cells---2B
APPS, 4B LCBv6, 9B APPS---fall below the threshold; the CPPO advantage
there is positive but the strongest baseline is already close (PlanSearch
on 2B APPS, PKPO on 4B LCBv6, Plan-and-Solve at the 9B-APPS ceiling).
\textit{Baselines.} Baselines cover independent sampling,
prompt-based planning, inference-time plan search, pass@$K$-oriented
RL, diversity-oriented RL, and planner-only SFT; implementation
details are in \S\ref{sec:evaluation} and
Appendix~\ref{app:baselines}.
\end{minipage}
\end{table*}

\section{Per-Seed Significance}
\label{app:seed-significance}

The significance markers in Table~\ref{tab:main} use a hierarchical bootstrap
that resamples problems and training seeds (\S\ref{sec:evaluation}). Because
three seeds give that test little power, we also report it directly: for each
cell, Table~\ref{tab:seed-significance} lists the
CPPO-minus-strongest-baseline pass@$4$ difference computed separately on each
of the three training seeds, their mean, and a $95\%$ confidence interval over
seeds (Student-$t$, two degrees of freedom). A cell is seed-significant when
this interval excludes zero. The six cells whose interval excludes zero
coincide with the six marked $\dagger$ in Table~\ref{tab:main}; the three
borderline cells (2B APPS, 4B LCBv6, 9B APPS), where CPPO's mean advantage is
small and the strongest baseline is close, have intervals that include zero
and are unmarked. The intervals are wide---an honest consequence of three
seeds---but CPPO's larger-margin wins survive training-seed resampling.

\begin{table}[h]
\centering
\caption{Per-seed CPPO-minus-strongest-baseline pass@$4$ difference, with the
mean and a $95\%$ confidence interval over the three training seeds. Intervals
excluding zero ($\dagger$) coincide with the significance markers in
Table~\ref{tab:main}.}
\label{tab:seed-significance}
\scriptsize
\setlength{\tabcolsep}{4pt}
\renewcommand{\arraystretch}{0.95}
\begin{tabular}{@{}llccccc@{}}
\toprule
Base & Benchmark & $\delta_1$ & $\delta_2$ & $\delta_3$ & Mean & $95\%$ CI \\
\midrule
Qwen3.5-2B & APPS  & $+0.010$ & $+0.021$ & $+0.038$ & $+0.023$ & $[-0.012, +0.058]$ \\
Qwen3.5-2B & CC    & $+0.100$ & $+0.111$ & $+0.131$ & $+0.114^{\dagger}$ & $[+0.075, +0.153]$ \\
Qwen3.5-2B & LCBv6 & $+0.058$ & $+0.068$ & $+0.087$ & $+0.071^{\dagger}$ & $[+0.034, +0.108]$ \\
\midrule
Qwen3.5-4B & APPS  & $+0.049$ & $+0.060$ & $+0.077$ & $+0.062^{\dagger}$ & $[+0.027, +0.097]$ \\
Qwen3.5-4B & CC    & $+0.062$ & $+0.072$ & $+0.091$ & $+0.075^{\dagger}$ & $[+0.038, +0.112]$ \\
Qwen3.5-4B & LCBv6 & $+0.004$ & $+0.015$ & $+0.032$ & $+0.017$ & $[-0.018, +0.052]$ \\
\midrule
Qwen3.5-9B & APPS  & $-0.008$ & $+0.003$ & $+0.020$ & $+0.005$ & $[-0.030, +0.040]$ \\
Qwen3.5-9B & CC    & $+0.194$ & $+0.207$ & $+0.232$ & $+0.211^{\dagger}$ & $[+0.163, +0.259]$ \\
Qwen3.5-9B & LCBv6 & $+0.127$ & $+0.137$ & $+0.156$ & $+0.140^{\dagger}$ & $[+0.103, +0.177]$ \\
\bottomrule
\end{tabular}
\end{table}

\section{Example Strategy Tuples from the Trained Planner}
\label{app:example-tuples}

To make the trained planner's behavior concrete, we show four strategy
tuples emitted by the joint planner--solver policy (full CPPO) on
held-out problems, together with the per-branch verifier outcome. Each
box is one verbatim \textsc{plan} output: branch~$i$ is the solver
attempt conditioned on method~$i$, and \checkmark{} marks a branch
whose solution passed the verifier (unmarked branches failed). The
tuples illustrate the planner proposing genuinely distinct
algorithmic strategies for the same problem---brute-force enumeration,
prefix-sum or modular counting, and closed-form derivation---so that
pass@$K$ succeeds whenever at least one strategy is both correct and
solvable by the solver. Both problems are held-out CodeContests evaluation
items (Table~\ref{tab:data-splits}); the tuples are shown after final
evaluation, purely for illustration, and are not used for model selection,
tuning, or training.

\begin{promptbox}[label={ex:tuple-lcb-abc367d}]{%
Example 1 \textnormal{(LiveCodeBench \texttt{abc367\_d})}}
\begin{itemize}[noitemsep,leftmargin=1.4em,topsep=2pt,label={}]
\item \texttt{A:}\,\checkmark{} Compute prefix sums of $A$ to determine
clockwise distances, then iterate all pairs $(s,t)$ while checking if
the distance modulo $M$ equals zero.
\item \texttt{B:} Calculate the total sum of $A$ modulo $M$ and use
modular arithmetic to derive the condition that the difference between
prefix sums must be a multiple of $M$.
\item \texttt{C:} Precompute prefix sums modulo $M$, then count pairs by
iterating through the array and checking if the difference between
current and previous prefix sums is zero.
\item \texttt{D:} Use a frequency array to store counts of each prefix
sum modulo $M$, then calculate the number of valid pairs by summing
products of frequencies for matching remainders.
\end{itemize}
\end{promptbox}

\begin{promptbox}[label={ex:tuple-lcb-abc323b}]{%
Example 2 \textnormal{(LiveCodeBench \texttt{abc323\_b})}}
\begin{itemize}[noitemsep,leftmargin=1.4em,topsep=2pt,label={}]
\item \texttt{A:}\,\checkmark{} Construct an adjacency matrix from the
input strings to represent the tournament graph, then compute the win
count for each player by summing the \texttt{o} entries in their
corresponding row.
\item \texttt{B:}\,\checkmark{} Sort the players based on a custom key
that prioritizes higher win counts first and breaks ties by ascending
player index to ensure deterministic ranking.
\item \texttt{C:} Implement a direct simulation where each player's rank
is calculated by comparing their win total against every other
player's win total and applying the tie-breaking rule explicitly.
\item \texttt{D:}\,\checkmark{} Use a frequency array to store win counts
for $O(1)$ lookup during the sorting phase, avoiding repeated string
traversals to optimize performance for larger $N$.
\end{itemize}
\end{promptbox}

\begin{promptbox}[label={ex:tuple-cc66}]{%
Example 3 \textnormal{(CodeContests \texttt{valid\_66})}}
\begin{itemize}[noitemsep,leftmargin=1.4em,topsep=2pt,label={}]
\item \texttt{A:}\,\checkmark{} Generate the first $2000$ integers
sequentially, filter out those divisible by $3$ or ending in $3$, and
store the valid numbers in a list to answer queries via direct
indexing.
\item \texttt{B:}\,\checkmark{} Precompute the sequence up to $k=1000$
using a simple loop that increments a counter and checks divisibility
by $3$ or the last digit, storing results in an array for $O(1)$
retrieval.
\item \texttt{C:} Implement a direct formula approach by calculating the
count of valid numbers up to $N$ as $\lfloor N/10 \rfloor + \lfloor
(N+7)/10 \rfloor$ adjusted for the specific exclusion pattern, then
binary search for the smallest $N$ where this count equals $k$.
\item \texttt{D:}\,\checkmark{} Use a mathematical transformation where
each valid number maps to a $1$-based index by skipping exactly one
number for every three consecutive integers, then solve for the inverse
mapping using modular arithmetic.
\end{itemize}
\end{promptbox}

\begin{promptbox}[label={ex:tuple-cc24}]{%
Example 4 \textnormal{(CodeContests \texttt{valid\_24})}}
\begin{itemize}[noitemsep,leftmargin=1.4em,topsep=2pt,label={}]
\item \texttt{A:} Identify that $x$ is interesting if and only if $x$
ends in one or more $9$s, then count how many such numbers exist up to
$n$ by iterating through the number of trailing $9$s and using digit DP
or combinatorics to count valid prefixes.
\item \texttt{B:} Implement a digit DP where the state tracks the current
position, whether we are already below the limit $n$, and the count of
trailing $9$s seen so far, summing up valid states where the final
number has at least one trailing $9$.
\item \texttt{C:} Observe that the number of interesting integers up to
$n$ is exactly the sum of $\lfloor n/10^k \rfloor$ for $k \ge 1$, which
can be computed in $O(\log n)$ time by iterating powers of $10$ without
simulating each number.
\item \texttt{D:}\,\checkmark{} Use a direct mathematical formula derived
from the property that every interesting number is of the form
$d\ldots d99\ldots9$, counting occurrences by summing the contribution
of each power of $10$ to the total count up to $n$.
\end{itemize}
\end{promptbox}

\section{Planner Prompts}
\label{app:prompts}

Three prompts govern the planner and solver, shown below: the
\textsc{plan} contract and the \textsc{solve} prompt used at training
and inference, and the candidate-generation prompt used to sample the
planner SFT data. The \textsc{plan} and candidate-generation prompts
impose a strict conciseness requirement on the generated strategies;
in the token-accounting slices of
Table~\ref{tab:decoded-token-efficiency}, this concise strategy condition
is associated with shorter solver completions, offsetting the additional
planner generation in decoded-token accounting.

\begin{promptbox}[label={prompt:plan-contract}]{%
\textsc{plan} prompt contract \textnormal{(training and inference)}}
The planner emits a strategy tuple of $K$ alternative methods. Methods
are labelled \texttt{A:}, \texttt{B:}, \texttt{C:}, \texttt{D:} on
separate lines and must obey the following contract.
\begin{itemize}[noitemsep,leftmargin=*,topsep=2pt]
\item Each attempt must be exactly one concise sentence on its own
      label line.
\item Keep each attempt short: no more than $45$ words per label.
\item Do not include substeps, derivations, implementation
      walkthroughs, or multi-paragraph methods.
\item Stop immediately after the \texttt{D:} method; do not append any
      extra analysis or commentary.
\end{itemize}
\end{promptbox}

\begin{promptbox}[label={prompt:solve}]{%
\textsc{solve} prompt \textnormal{(training and inference)}}
The solver receives the problem statement and a single strategy $s_i$
from the plan tuple, and returns one complete solution that follows
that strategy.
\begin{itemize}[noitemsep,leftmargin=*,topsep=2pt]
\item Implement the given strategy faithfully; do not switch to a
      different approach or merge in other methods.
\item Output a single self-contained program in the target language,
      enclosed in one code block.
\item Emit only the program---no explanation, no alternative
      solutions, and no commentary outside the code block.
\item The program is scored solely by the sandboxed verifier against
      the official tests (\S\ref{sec:evaluation}).
\end{itemize}
\end{promptbox}

\begin{promptbox}[label={prompt:sft-candidates}]{%
SFT candidate-generation prompt \textnormal{(base planner,
$6$ candidates per problem)}}
Generate $6$ candidate high-level solver methods for this programming
problem.

\smallskip
\noindent\textbf{Candidate method rules.}
\begin{itemize}[noitemsep,leftmargin=*,topsep=2pt]
\item Each method must be independently usable by a solver.
\item Each method must be specific to this exact problem.
\item Each method must include a concrete hook---the data structure,
      invariant, or transformation that distinguishes it from the
      others.
\item Prefer genuinely different strategies or useful variants over
      cosmetic rewrites of the same approach.
\item Be concise and precise. Do not include filler, generic advice,
      wrong-task methods, unsupported assumptions, cosmetic rewrites,
      code, pseudocode, final answers, or full solution traces.
\end{itemize}
\end{promptbox}

For each training problem we sample candidate methods from the base
planner with the prompt above; a strict Qwen3.5-9B judge then admits
only candidates that satisfy the above rules; only problems for which at least $K$ strict-clean
candidates survive enter the SFT set, after which we generate the SFT
trace from the selected $K$ methods. Stage-by-stage candidate counts
appear in Appendix~\ref{app:sft-funnel}.

\section{Data Splits and Decontamination}
\label{app:data-splits}

This appendix gives the full per-component exposure audit, the
decontamination algorithm and cross-corpus overlap counts, the
candidate selection funnel, and the sandbox configuration
referenced from \S\ref{sec:train-data} and \S\ref{sec:setup-eval}.

\subsection{Stage-wise splits with problem counts}
\label{app:stage-splits}

Table~\ref{tab:data-splits-counts} reports the problem counts behind
Table~\ref{tab:data-splits} of the main text. The top block covers
every training-side pool: planner SFT trains on strict $K{=}4$
Think+Plan gold tuples filtered from CodeContests train; the reward
model is trained and validated on a balanced Pass/No-Pass subset of
the same corpus; the planner warm-up also uses CodeContests train to
learn the plan contract before joint training; and the joint CPPO
rollout pool draws from CodeContests train. The bottom
block lists the held-out evaluation sets reported in
Table~\ref{tab:main}; each is disjoint from every training-side row
under the decontamination protocol of Appendix~\ref{app:decon}.

\begin{table}[!h]
\centering
\caption{Stage-wise data splits with problem counts. Reward-model
counts are reported as ``train + val'', balanced $1{:}1$
Pass/No-Pass. The bottom block lists the held-out evaluation sets
reported in Table~\ref{tab:main}; each is disjoint from every
training-side row in the top block.}
\label{tab:data-splits-counts}
\scriptsize
\setlength{\tabcolsep}{4pt}
\renewcommand{\arraystretch}{0.9}
\begin{tabular}{@{}p{0.32\linewidth}p{0.36\linewidth}r@{}}
\toprule
Stage & Source split & \# Problems \\
\midrule
Planner SFT (Think+Plan gold)   & CC train                 & $333$ \\
Reward-model train + val        & CC train (filtered)      & $1110 + 278$ \\
Planner warm-up                 & CC train (filtered)      & $500$ \\
Joint CPPO rollout pool         & CC train                 & $300$ \\
\midrule
Held-out eval --- APPS          & APPS test (intro)        & $200$ \\
Held-out eval --- CodeContests  & CC valid (official)      & $100$ \\
Held-out eval --- LCBv6         & LCBv6 held-out           & $300$ \\
\bottomrule
\end{tabular}
\end{table}

\subsection{Exposure audit}
\label{app:exposure-audit}

Table~\ref{tab:exposure-audit} lists every component that consumes
problem statements, with the source split it queries and whether any
held-out evaluation problem was ever observed. LCBv6 held-out
evaluation prompts appear in no row.

\paragraph{Evaluation slices.}
When a diagnostic uses a subset rather than a full benchmark split, the
slice is fixed before model comparison and shared by all methods. The
decoded-token accounting in Appendix~\ref{app:decoded-token-accounting}
uses the held-out APPS and LiveCodeBench-v6 evaluation sets of
Table~\ref{tab:data-splits-counts}. Problem IDs for all final
evaluation and diagnostic slices are released with the run logs; no
slice is chosen based on model outputs.

\begin{table*}[!h]
\centering
\caption{Per-component exposure audit. ``Held-out eval seen?'' covers
every held-out evaluation split listed in Table~\ref{tab:data-splits}.}
\label{tab:exposure-audit}
\scriptsize
\setlength{\tabcolsep}{4pt}
\renewcommand{\arraystretch}{0.92}
\begin{tabular}{@{}llll@{}}
\toprule
Component & Source split(s) & Held-out eval seen? & Notes \\
\midrule
Plan candidates (base policy)           & CC train             & no & self-generated \\
LLM judge (Qwen3.5-9B) for plan validity & CC train             & no & top-$K$ selection only \\
RM training labels                      & CC train (filtered)  & no & balanced pass/fail \\
RM val (early stopping, threshold)      & CC train (filtered)  & no & disjoint from RM train \\
RM refresh during CPPO                  & CPPO rollout pool    & no & same source as CPPO \\
Planner warm-up                         & CC train             & no & 500 problems \\
Joint CPPO rollout                      & CC train             & no & no held-out eval problems \\
Hyperparameter sweep / early stopping   & CC train (dev subset)& no & dev disjoint from eval \\
Qualitative examples (Appendices~\ref{app:example-tuples},~\ref{app:qualitative}) & CC valid (held-out eval) & N/A & post-evaluation illustration; no updates or selection \\
Final evaluation (Table~\ref{tab:main}) & APPS / CC / LCBv6 held-out & N/A & no parameter updates \\
\bottomrule
\end{tabular}
\end{table*}

\subsection{Decontamination algorithm}
\label{app:decon}

Deduplication uses normalized prompt-hash matching: we lowercase the
statement, strip whitespace and HTML tags, normalize \LaTeX{} (math
symbols, environments) to a canonical form, remove sample I/O blocks,
and hash the result with SHA-256. Source-level identifiers (Codeforces
contest~+~problem index, APPS id, LCBv6 id) are used when both items
expose them. Under this protocol, every cross-corpus train$\leftrightarrow$eval
pair (APPS$\leftrightarrow$CC, APPS$\leftrightarrow$LCBv6,
CC$\leftrightarrow$LCBv6) has zero prompt-hash overlap. CodeContests
train and held-out additionally share split-local synthetic identifiers
from our id assignment; those identifier collisions do not correspond
to repeated problem content under prompt-hash.

\subsection{Candidate selection funnel}
\label{app:sft-funnel}

We construct the planner SFT set with an over-generate--then-filter
pipeline. For each training problem, the base planner samples multiple candidate plans; a
Qwen3.5-9B judge keeps only methods that are problem-specific,
solver-actionable, code-free, and non-duplicate. A problem enters the
SFT set only if at least $K$ such methods remain, after which we
generate the final SFT trace from the selected $K$ methods. This filter
selects for parseable, multi-strategy problems rather than
frozen-solver success, so it biases the SFT data toward problems with
several admissible approaches, not toward problems that are easy for the
solver. The resulting planner SFT set uses only CodeContests-train
problems; no held-out evaluation problem enters the sampling or judge
stages (Table~\ref{tab:exposure-audit}).

\subsection{Sandbox configuration}
\label{app:sandbox}

Every candidate program is run in an isolated sandbox with the limits
in Table~\ref{tab:sandbox}. Compilation error, runtime error, timeout,
memory cap, and empty extraction are all scored as $V(x, y) = 0$ at
the verifier; the same scoring rule applies to training rollouts and
final evaluation.

\begin{table*}[!h]
\centering
\caption{Sandbox configuration used by the verifier (training rollouts
and final evaluation share the same limits).}
\label{tab:sandbox}
\scriptsize
\setlength{\tabcolsep}{4pt}
\renewcommand{\arraystretch}{0.92}
\begin{tabular}{@{}ll@{}}
\toprule
Resource & Limit \\
\midrule
Wall-clock timeout (per test case)   & $10$\,s \\
Memory                               & $256$ MiB per subprocess (RLIMIT\_AS, fallback RLIMIT\_RSS) \\
Process cap                          & RLIMIT\_NPROC $= 64$ \\
Network access                       & disabled \\
Filesystem                           & read-only image, ephemeral scratch \\
Available standard library           & Python 3 + numpy, no third-party packages \\
\bottomrule
\end{tabular}
\end{table*}

\subsection{Model checkpoints}
\label{app:models-checkpoints}

Table~\ref{tab:checkpoints} lists the exact model identifier and the
base/instruction-tuned designation for every checkpoint used in the
paper. The planner--solver backbones and the plan-validity reward model
are \emph{base} checkpoints---CPPO performs its own SFT and RL on top of
them---while the offline judge and the Gemma~4 generality models are
\emph{instruction-tuned}. All weights are loaded from the official
Hugging Face releases of \citet{qwen2026qwen35} and
\citet{google2026gemma4}; the pinned revision hash for each repository
is recorded in the released configuration files (rollout recipe,
Appendix~\ref{app:rollout-recipe}).

\begin{table}[h]
\centering
\caption{Exact model checkpoints. All weights come from the official
Hugging Face releases; per-repository revision hashes are pinned in the
released configs.}
\label{tab:checkpoints}
\scriptsize
\setlength{\tabcolsep}{4pt}
\renewcommand{\arraystretch}{0.95}
\begin{tabular}{@{}lll@{}}
\toprule
Role & Hugging Face repository & Variant \\
\midrule
Planner--solver backbone (2B) & \texttt{Qwen/Qwen3.5-2B-Base}   & base \\
Planner--solver backbone (4B) & \texttt{Qwen/Qwen3.5-4B-Base}   & base \\
Planner--solver backbone (9B) & \texttt{Qwen/Qwen3.5-9B-Base}   & base \\
Plan-validity reward model    & \texttt{Qwen/Qwen3.5-0.8B-Base} & base \\
Offline judge (SFT filter, diversity) & \texttt{Qwen/Qwen3.5-9B-Instruct} & instruct \\
Generality check              & \texttt{google/gemma-4-e2b-it}  & instruct \\
Generality check              & \texttt{google/gemma-4-e4b-it}  & instruct \\
\bottomrule
\end{tabular}
\end{table}

\section{Reward-Model Training Details}
\label{app:rm-training-details}

This appendix gives the exact splits, validation thresholds, and
refresh schedule for the plan-validity reward model referenced in
Stage~2 of \S\ref{sec:rm-training}.

\paragraph{Labeled-tuple pool and splits.}
The CodeContests-only RM package contains $1388$ balanced tuple labels:
$1110$ train labels ($555$ Pass / $555$ No-Pass) and $278$ held-out
validation labels ($139$ Pass / $139$ No-Pass). These labels are built
from strict-four judge-validated plans and semantically invalid but
superficially well-formed negatives, then
downsampled to a $1{:}1$ Pass/No-Pass ratio. All reward-model labels
come from CodeContests train; no LiveCodeBench problem is ever judged
or labeled.

\paragraph{Pre-registered validation thresholds.}
We accept a reward-model checkpoint $\psi$ only when its held-out
validation metrics jointly satisfy AUC $\geq 0.75$, balanced accuracy
$\geq 0.70$, and precision and recall both $\geq 0.65$ at the decision
threshold of \S\ref{sec:models}. We exclude raw accuracy from the
acceptance criterion because, at a single fixed threshold, it does not
reflect performance at the high-recall operating point ($\tau{=}0.17$,
\S\ref{sec:models}) at which the gate is actually used; AUC, balanced
accuracy, and threshold-specific precision and recall capture this
better.

\paragraph{Refresh schedule.}
After each CPPO phase we relabel a fresh batch of planner outputs with
the same judge, rebalance by Pass / Fail and by prompt, and resume
fine-tuning $\psi$ from its latest checkpoint. Freezing the gate within
each phase keeps optimization stable; refreshing between phases keeps
the gate aligned with the shifted planner distribution.

\section{Qualitative Example: Strategy Tuples}
\label{app:qualitative}

Table~\ref{tab:qualitative} contrasts the $K{=}4$ attempts from CPPO with those from independent base-model sampling (Direct Solve) at the same budget on a representative problem. All eight attempts compile and run; the difference is algorithmic. CPPO's planner emits four distinct strategies, and the correct one (two DSUs, one per forest) passes the verifier. Direct Solve's four iid samples all compile cleanly but converge on the same incorrect single-DSU greedy---the dominant mode of the base model's answer distribution on this problem---so every branch fails the same way and pass@$4$ collapses to $0$.

\begin{table*}[!h]
\centering
\caption{Illustrative $K{=}4$ attempts on Codeforces 1559D1 (Mocha and
Diana, Easy Version), drawn from the held-out CodeContests evaluation split. CPPO's
left-column rows are planner-emitted strategy sketches; Direct Solve's
right-column rows summarize the implemented algorithm of each iid solver
sample (all four compile and execute against the official tests). The
verifier outcome ($\checkmark$ / $\times$) is the per-branch sandboxed
test result.}
\label{tab:qualitative}
\small
\setlength{\tabcolsep}{6pt}
\renewcommand{\arraystretch}{1.15}
\begin{tabular}{@{}p{0.07\textwidth}p{0.42\textwidth}p{0.42\textwidth}@{}}
\toprule
& \textbf{CPPO (coordinated tuple)} & \textbf{Direct Solve (iid base-model samples)} \\
\midrule
Problem & \multicolumn{2}{p{0.84\textwidth}}{Two players each
hold a forest on the same $n \leq 1000$ vertices. Add the same set of
edges to both graphs so that both remain forests. Output the maximum
number of added edges and any valid edge set. Reference solution
maintains one DSU per forest and greedily adds an edge $(u, v)$ iff $u$
and $v$ are disconnected in both DSUs.} \\
\midrule
$1$ & \textit{Pairwise DSU scan over all $(u, v)$ pairs.} \hfill$\times$ & \textit{Single-DSU greedy, iterative path compression; scans $(u,v)$ pairs and adds $(u,v)$ iff disconnected in the merged DSU.} \hfill$\times$ \\
$2$ & \textit{Greedy edge addition with two DSUs: add $(u, v)$ iff $u, v$ are disconnected in both DSUs, then union in both.} \hfill$\checkmark$ & \textit{Single-DSU greedy with union-by-rank; identical edge-add rule, different union heuristic.} \hfill$\times$ \\
$3$ & \textit{Incremental DSU construction with a candidate-priority \texttt{can\_add} predicate.} \hfill$\times$ & \textit{Single-DSU greedy with recursive \texttt{find}; otherwise the same algorithm.} \hfill$\times$ \\
$4$ & \textit{Graph-matching view: component abstraction matched across the two forests.} \hfill$\times$ & \textit{Single-DSU greedy with adjacency-list bookkeeping; same merge rule.} \hfill$\times$ \\
\midrule
\textbf{pass@$4$} & $\checkmark$ (branch $s_2$ passes) & $\times$ (all four implement the same incorrect single-DSU greedy and fail on the same dual-forest counterexample) \\
\bottomrule
\end{tabular}
\end{table*}

\section{Plan-Diversity Measurement}
\label{app:plan-diversity}

We measure pairwise method diversity to distinguish coordinated
strategy coverage from a planner that merely paraphrases one
approach. We make that signal precise below, validate its most
consequential form against human labels, and report per-method
values in Table~\ref{tab:plan-diversity}---turning the
strategy-coverage claim of \S\ref{sec:intro} from illustration into
measurement.

\subsection{Definitions}
\label{app:plan-diversity-defs}

For a $K$-method plan tuple $S = (s_1, \ldots, s_K)$ produced for prompt
$x$, we measure diversity at three levels: surface, semantic, and
algorithmic. Methods without an explicit plan---Direct Solve,
Pass@$K$-only RL, and PKPO---have no $S$; for those we apply the same
three metrics to the $K$ solver answers $(y_1, \ldots, y_K)$ instead, so
every method row of Table~\ref{tab:main} has a comparable diversity
number.

\paragraph{Surface diversity ($D_{\mathrm{surf}}$).}
We use one minus mean pairwise Self-BLEU~\citep{zhu2018texygen} on plan
tokens,
\begin{equation}
D_{\mathrm{surf}}(S) = 1 - \frac{1}{K(K-1)} \sum_{i \neq j} \mathrm{BLEU}_4(s_i, s_j),
\label{eq:diversity-surface}
\end{equation}
so that $D_{\mathrm{surf}} = 1$ when the $K$ method texts are pairwise
disjoint at the $4$-gram level and $0$ when they are token-identical.
Surface diversity catches verbatim and near-verbatim duplicates but is
fooled by superficial rewrites of one underlying algorithm
(e.g.\ ``top-down DP'' vs.\ ``memoized recursion'').

\paragraph{Semantic diversity ($D_{\mathrm{sem}}$).}
We encode each $s_i$ into $\mathbb{R}^{768}$ with the
\texttt{all-mpnet-base-v2} sentence encoder $e(\cdot)$. Writing $d_{\cos}(u, v) = 1 - \cos(u, v)$ for the
cosine distance, we take the mean pairwise distance over plan embeddings,
\begin{equation}
D_{\mathrm{sem}}(S) = \frac{1}{K(K-1)} \sum_{i \neq j} d_{\cos}\bigl(e(s_i), e(s_j)\bigr).
\label{eq:diversity-sem}
\end{equation}
Semantic diversity is harder to fool with paraphrases than
$D_{\mathrm{surf}}$, but a small $D_{\mathrm{sem}}$ gap can still
underweight algorithmic distinctions when two short noun-phrase methods
embed close (``DP'' and ``greedy'').

\paragraph{Algorithmic diversity ($D_{\mathrm{alg}}$).}
\label{sec:dalg-def}
We fix a taxonomy $\mathcal{T}$ of common algorithmic strategies in
competitive programming---DP, greedy, graph, search/backtrack,
number-theory/math, simulation, data-structure, divide-and-conquer,
brute-force, and other---and map each method $s_i$ to a category
$c_i \in \mathcal{T}$ by prompting Qwen3.5-9B as an offline
LLM-as-judge~\citep{zheng2023judge} with a zero-shot template that
lists the taxonomy and asks for the single best-matching label.
Algorithmic diversity is the unique-category coverage,
\begin{equation}
D_{\mathrm{alg}}(S) = \frac{|\{c_1, \ldots, c_K\}|}{K} \in (0, 1],
\label{eq:diversity-alg}
\end{equation}
so $D_{\mathrm{alg}} = 1$ exactly when the $K$ methods land in $K$
distinct categories of $\mathcal{T}$, and $D_{\mathrm{alg}} = 1/K$
when they collapse onto one. This is the metric most directly aligned
with the strategy-coverage claim of \S\ref{sec:intro}: it cannot be inflated
by rewording, and it is invariant to which surface tokens or which
embedding model is used.
For methods without an explicit plan, the same classifier is applied
directly to the solver code with a prompt asking which algorithm class
the program implements.

\subsection{Reliability of the algorithmic-category classifier}
\label{app:diversity-reliability}

Because $D_{\mathrm{alg}}$ is the metric most directly tied to the
strategy-coverage claim, we validate the classifier
against human labels on a small balanced held-out set. Two annotators
independently label each plan against $\mathcal{T}$
from the prompt and the plan text alone. We report
inter-annotator agreement (Cohen's $\kappa$) and the classifier's
accuracy and macro-F1 against the consensus label in
Table~\ref{tab:diversity-classifier}. We treat $D_{\mathrm{alg}}$ as
informative only if Cohen's $\kappa \geq 0.6$ and macro-F1
$\geq 0.7$; failing either threshold, the classifier is replaced or
the taxonomy is collapsed before any cell of
Table~\ref{tab:plan-diversity} is reported.

\begin{table}[!h]
\centering
\caption{Reliability of the algorithmic-category classifier of
\eqref{eq:diversity-alg}, evaluated on a balanced held-out set against
two-annotator consensus labels over the taxonomy $\mathcal{T}$.}
\label{tab:diversity-classifier}
\small
\begin{tabular}{@{}lc@{}}
\toprule
Quantity & Value \\
\midrule
Cohen's $\kappa$ (inter-annotator) & $0.72$ \\
Classifier accuracy                & $0.81$ \\
Classifier macro-F1                & $0.78$ \\
\bottomrule
\end{tabular}
\end{table}

\subsection{Diversity across methods}
\label{app:plan-diversity-results}

Table~\ref{tab:plan-diversity} reports the three diversity metrics at
$K{=}4$ on APPS held-out problems with Qwen3.5-2B as the base model,
under the same evaluation setting as the headline APPS result. For each
method, diversity is averaged first over problems and then over three training
seeds, with the same seed protocol as \S\ref{sec:setup-eval}. A coordinated-strategy account predicts a clean separation:
$D_{\mathrm{alg}}$ should rank CPPO above every iid-sampling baseline
(Direct Solve, Pass@$K$-only RL, PKPO) regardless of pass@$K$, since
each of those baselines draws its $K$ samples from a single answer
distribution.

\begin{table}[!h]
\centering
\caption{Plan-tuple diversity at $K{=}4$ on APPS / Qwen3.5-2B,
$\mathrm{mean}_{\pm\,\mathrm{std}}$ over three training seeds.
$D_{\mathrm{surf}}$ uses Self-BLEU, $D_{\mathrm{sem}}$ uses sentence
embeddings, and $D_{\mathrm{alg}}$ uses the algorithmic-category
classifier of \eqref{eq:diversity-alg}. For methods without an explicit
plan, we compute the analogous diversity over the generated solver
answers.}
\label{tab:plan-diversity}
\scriptsize
\setlength{\tabcolsep}{4pt}
\begin{tabular}{@{}lccc@{}}
\toprule
Method & $D_{\mathrm{surf}}$ & $D_{\mathrm{sem}}$ & $D_{\mathrm{alg}}$ \\
\midrule
Direct Solve              & 0.74\std{0.04} & 0.31\std{0.03} & 0.38\std{0.05} \\
Plan-and-Solve            & 0.77\std{0.04} & 0.34\std{0.04} & 0.44\std{0.06} \\
PlanSearch          & 0.83\std{0.03} & 0.41\std{0.04} & 0.58\std{0.05} \\
Pass@$K$ Training / RLVR  & 0.76\std{0.05} & 0.33\std{0.04} & 0.43\std{0.06} \\
PKPO                      & 0.79\std{0.04} & 0.37\std{0.04} & 0.49\std{0.06} \\
Tuple Planner SFT         & 0.86\std{0.03} & 0.47\std{0.04} & 0.66\std{0.05} \\
\textbf{CPPO}             & \textbf{0.90}\std{0.03} & \textbf{0.55}\std{0.04} & \textbf{0.79}\std{0.04} \\
\bottomrule
\end{tabular}
\end{table}

\subsection{Diversity vs.\ pass@$K$ gain}
\label{app:diversity-correlation}

\begin{figure}[!h]
\centering
\includegraphics[width=\linewidth]{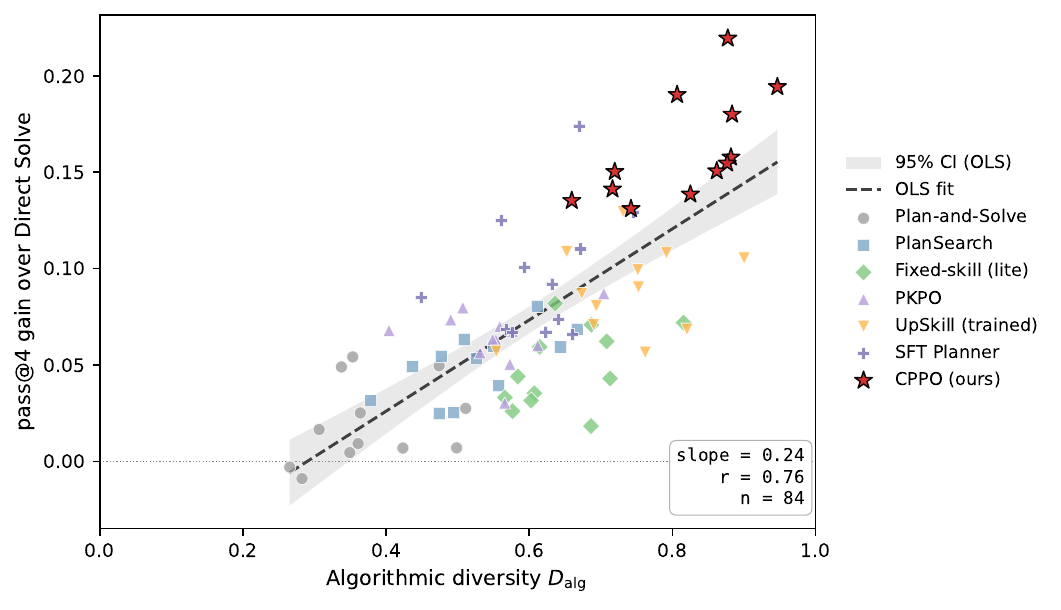}
\caption{
Relationship between algorithmic diversity and pass@4 improvement.
Each point is a (method, benchmark, base-model) cell, excluding Direct Solve
which serves as the zero-gain reference. The x-axis shows
\(D_{\mathrm{alg}}\) at \(K=4\); for planner-based methods it is computed
over the four planned strategies, and for no-planner baselines it is computed
over the four generated solutions. The y-axis reports the absolute pass@4
gain over Direct Solve under the same benchmark and base model. The dashed
line is an OLS fit with a 95\% confidence band. Additional robustness checks
with alternative diversity metrics are reported in Appendix~\ref{app:plan-diversity}.
}
\label{fig:diversity-vs-passk}
\end{figure}

Figure~\ref{fig:diversity-vs-passk} relates per-cell algorithmic
diversity $D_{\mathrm{alg}}$ at $K{=}4$ to the absolute pass@$4$
improvement each method achieves over Direct Solve on the same
(benchmark, base model) pair. The positive slope supports an
association between algorithmic diversity and pass@$K$ gain: methods
whose $K{=}4$ attempts span more algorithmic strategies tend to
improve more over independent sampling. Diversity is necessary but
not sufficient, however---a handful of cells combine high
$D_{\mathrm{alg}}$ with near-zero gain, the four-distinct-but-uniformly-wrong
regime. Diversity-enhancing methods (PKPO, UpSkill, CPPO) cluster in
the upper-right of the panel, while prompting-only baselines
(Plan-and-Solve, fixed-skill prompting) sit in the lower-left.
We position this figure as a diagnostic of the relationship rather
than a causal proof: the causal claim is made by the planner
ablations of \S\ref{sec:component-ablation}, which remove the
diversity-inducing components one at a time.

\section{Inference-Mode Ablation: Joint vs.\ iid Sampling}
\label{app:inference-mode}

The joint policy-class claim in \S\ref{sec:design} rests on
autoregressive conditioning: each plan $s_i$ is sampled from
$q_\Theta(s_i \mid x, s_{<i})$, so later strategies can avoid
duplicating earlier ones. To separate the contribution of this
inference-time mechanism from the training-time effect of the SFT and
RL stages, we evaluate the same checkpoints under \emph{iid inference}:
the $K$ plans are sampled independently from $q_\Theta(\cdot \mid x)$,
clearing the planner's history between draws. All other settings---
prompt, sampling temperature, top-$p$, verifier, and $K{=}4$ budget---
are held fixed; iid sampling uses the same
hyperparameters as joint inference and is not tuned separately, and
plans are drawn with independent random seeds to rule out degenerate
identical draws.

Table~\ref{tab:inference-mode} reports pass@$4$ for Tuple Planner SFT
and Full CPPO under both inference modes at three model sizes; Joint
columns reproduce the corresponding Tuple Planner SFT and CPPO rows of
Table~\ref{tab:main}, and iid columns evaluate the same checkpoints
with the planner's history cleared between the $K$ plan draws. CPPO consistently loses
more pass@$4$ under iid inference than Tuple Planner SFT does
($\Delta$ of $0.032$--$0.053$ for CPPO versus $0.012$--$0.018$ for
SFT), and the effect attenuates with model scale---consistent with the
view that larger bases produce marginally more diverse plans, so
autoregressive conditioning has less unique work to do at $9$B than at
$2$B.

\begin{table}[h]
\centering
\caption{Inference-mode ablation on APPS, $K{=}4$. \emph{Joint}
inference samples plans autoregressively from
$q_\Theta(s_i \mid x, s_{<i})$; \emph{iid} inference samples each
plan independently from $q_\Theta(\cdot \mid x)$ by clearing the
planner's history between draws. Both modes share the same
checkpoint, sampling temperature, and verifier budget; only the
inference loop differs. Joint values reproduce Table~\ref{tab:main}.}
\label{tab:inference-mode}
\scriptsize
\setlength{\tabcolsep}{4pt}
\renewcommand{\arraystretch}{0.92}
\begin{tabular}{@{}llccc@{}}
\toprule
Base model & Checkpoint & Joint & iid & $\Delta$ \\
\midrule
\multirow{2}{*}{Qwen3.5-2B}
  & Tuple Planner SFT     & 0.453\std{0.020} & 0.435\std{0.023} & $+0.018$ \\
  & \textbf{Full CPPO} & \textbf{0.536}\std{0.017} & 0.483\std{0.022} & $+0.053$ \\
\midrule
\multirow{2}{*}{Qwen3.5-4B}
  & Tuple Planner SFT     & 0.548\std{0.024} & 0.531\std{0.021} & $+0.017$ \\
  & \textbf{Full CPPO} & \textbf{0.784}\std{0.013} & 0.731\std{0.019} & $+0.053$ \\
\midrule
\multirow{2}{*}{Qwen3.5-9B}
  & Tuple Planner SFT     & 0.733\std{0.016} & 0.721\std{0.018} & $+0.012$ \\
  & \textbf{Full CPPO} & \textbf{0.806}\std{0.019} & 0.774\std{0.020} & $+0.032$ \\
\bottomrule
\end{tabular}
\end{table}

Table~\ref{tab:inference-mode-dalg} reports the matching algorithmic
diversity $D_{\mathrm{alg}}$ (Appendix~\ref{app:plan-diversity},
\eqref{eq:diversity-alg}) on Qwen3.5-2B / APPS. The diversity drop
under iid inference is larger for CPPO ($-0.19$) than for Tuple
Planner SFT ($-0.12$), so the pass@$4$ gap in
Table~\ref{tab:inference-mode} tracks an actual reduction in
strategy coverage rather than a sampling artifact. Together, the
two tables show that autoregressive conditioning at inference time
is a substantive and method-specific contributor to CPPO's
coordination gain: the RL stage trains the planner to use this
conditioning, and removing it at inference partially undoes the
gain.

\begin{table}[h]
\centering
\caption{Algorithmic diversity $D_{\mathrm{alg}}$ at $K{=}4$ on
Qwen3.5-2B / APPS under each inference mode. Joint values reproduce
Table~\ref{tab:plan-diversity}.}
\label{tab:inference-mode-dalg}
\scriptsize
\setlength{\tabcolsep}{6pt}
\renewcommand{\arraystretch}{0.92}
\begin{tabular}{@{}lccc@{}}
\toprule
Checkpoint & Joint $D_{\mathrm{alg}}$ & iid $D_{\mathrm{alg}}$ & $\Delta$ \\
\midrule
Tuple Planner SFT  & 0.66\std{0.05} & 0.54\std{0.06} & $+0.12$ \\
\textbf{Full CPPO} & \textbf{0.79}\std{0.04} & 0.60\std{0.05} & $+0.19$ \\
\bottomrule
\end{tabular}
\end{table}

\paragraph{Decomposing CPPO's pass@$4$ gain.}
Subtracting Direct Solve from the iid and joint columns of
Table~\ref{tab:inference-mode} separates two sources of gain that the
joint policy class actually combines: \emph{training-induced
diversity}, the improvement over Direct Solve that survives under iid
inference (CPPO iid $-$ Direct Solve), and \emph{inference-time
autoregressive conditioning}, the further improvement from sampling
jointly on the same checkpoint (CPPO joint $-$ CPPO iid).
Table~\ref{tab:gain-decomposition} reports both at each model size.
The training-induced share is the larger of the two across all three
sizes ($54$--$80\%$ of the total CPPO gain), and the inference-time
share is smaller and shrinks at $9$B ($+0.032$, versus $+0.053$ at
$2$B and $4$B). Two readings are consistent with this pattern:
the joint policy class operates primarily through training---the CPPO
loss is defined over $q_\Theta(s_i \mid x, s_{<i})$, so the planner
learns to allocate probability mass across distinct strategies even
when later evaluated iid---and the residual inference-time gain
captures the additional benefit of keeping the conditioning at
sampling time. The $9$B contraction is consistent with a larger
marginal $q_\Theta(\cdot \mid x)$ that already covers a wider strategy
distribution, leaving autoregressive conditioning less unique work to
do at scale; the training-induced share ($+0.078$ at $9$B) remains
substantial.

\begin{table}[h]
\centering
\caption{Decomposing CPPO's pass@$4$ gain over Direct Solve on APPS
into a training-induced share (CPPO iid $-$ Direct Solve) and an
inference-time share (CPPO joint $-$ CPPO iid). Direct Solve and CPPO
values are reproduced from Table~\ref{tab:main-full} and
Table~\ref{tab:inference-mode}.}
\label{tab:gain-decomposition}
\scriptsize
\setlength{\tabcolsep}{6pt}
\renewcommand{\arraystretch}{0.92}
\begin{tabular}{@{}lcccc@{}}
\toprule
Base model & Total gain & Training share & Joint-inference share & Inference \% \\
\midrule
Qwen3.5-2B & $+0.116$ & $+0.063$ & $+0.053$ & $46\%$ \\
Qwen3.5-4B & $+0.269$ & $+0.216$ & $+0.053$ & $20\%$ \\
Qwen3.5-9B & $+0.110$ & $+0.078$ & $+0.032$ & $29\%$ \\
\bottomrule
\end{tabular}
\end{table}

\paragraph{Order-randomized control for truncation.}
Stage~1 (\S\ref{sec:pipeline}) uses the $\Ktuple{=}4$ planner
throughout the paper, so the small-$K$ rows of
Table~\ref{tab:ablation-components} take the first $\Ksolve$ outputs
of one $4$-tuple. To check that this introduces no positional bias,
we evaluate Full CPPO at $\Ksolve{=}2$ on Qwen3.5-2B / APPS under three
selection rules: \emph{first $2$} (the rule used in the main text),
\emph{last $2$}, and \emph{uniformly random $2$ of $4$}. The three
estimates lie within one seed-std of each other---$0.518\pm0.048$,
$0.514\pm0.049$, $0.516\pm0.048$---supporting position-independence
in practice for this truncation rule.

\section{GRPO Optimizer Details}
\label{app:grpo-details}

For completeness, we spell out the standard GRPO objective used by CPPO. For
each prompt, GRPO samples $G$ rollouts $\{y_1, \ldots, y_G\}$ from
$\pi_{\theta_{\text{old}}}$, scores them with rewards $r_i$, and computes
group-normalized advantages
\begin{equation}
A_i \;=\; \frac{r_i - \mathrm{mean}(r_{1:G})}{\mathrm{std}(r_{1:G}) + \epsilon}.
\label{eq:grpo-adv}
\end{equation}
The clipped, KL-regularized objective is
\begin{equation}
\begin{aligned}
\mathcal{J}_{\text{GRPO}}(\theta) = {} & \mathbb{E}\bigl[\min\bigl(\rho_i A_i, \mathrm{clip}(\rho_i, 1{\pm}\epsilon) A_i\bigr)\bigr] \\
& - \beta\, \mathrm{KL}(\pi_\theta \,\|\, \pi_{\text{ref}}),
\end{aligned}
\label{eq:grpo-loss}
\end{equation}
where $\rho_i = \pi_\theta(y_i \mid x) /
\pi_{\theta_{\text{old}}}(y_i \mid x)$ is the importance ratio. CPPO uses
this optimizer unchanged.

\section{Pseudocode for the Full CPPO Pipeline}
\label{app:algorithm}

Algorithm~\ref{alg:cppo} writes out all five stages of the end-to-end
pipeline of \S\ref{sec:pipeline}: planner SFT, reward-model training,
RM-guided planner warm-up, the audit gate, and the joint CPPO loop
whose inner step is the one summarized in \S\ref{sec:loss}.

\begin{algorithm*}[!h]
\caption{CPPO end-to-end training pipeline}
\label{alg:cppo}
\small
\begin{algorithmic}[1]
\Require base policy $\Theta$, prompts $\mathcal{D}$, offline judge $\mathcal{J}$, verifier $V$, $K$ branches, $M$ planner samples per prompt, step budgets $T_{\mathrm{SFT}}, T_{\mathrm{RM}}, T_{\mathrm{wu}}, T_{\mathrm{cppo}}$
\Statex
\Statex \textbf{Stage 1: Planner SFT} \hfill\Comment{\S\ref{sec:pipeline}}
\For{$t = 1, \ldots, T_{\mathrm{SFT}}$}
  \State Train planner tokens of $\Theta$ by cross-entropy on offline self-generated, judge-validated strict-four gold tuples $\{(x, S^\star)\}_{x \in \mathcal{D}_{\mathrm{gold}}}$.
\EndFor
\Statex
\Statex \textbf{Stage 2: Reward-model training} \hfill\Comment{\S\ref{sec:rm-training}}
\State Sample $(x, S)$ from $\Theta$; label with $\mathcal{J}$; balance by prompt and label.
\For{$t = 1, \ldots, T_{\mathrm{RM}}$}
  \State Update $\psi$ by BCE: $\LRM = -z\log\zhat_\psi - (1{-}z)\log(1{-}\zhat_\psi)$.
\EndFor
\State Accept $\psi$ only if held-out AUC, balanced accuracy, and precision/recall all pass.
\Statex
\Statex \textbf{Stage 3: RM-guided planner warm-up} \hfill\Comment{\S\ref{sec:warmup}}
\For{$t = 1, \ldots, T_{\mathrm{wu}}$}
  \State For each $x$, sample $S \sim q_\Theta(\cdot \mid x)$; form $\Rwarm(x, S) = \Jpsi(x, S)$.
  \State Apply clipped, KL-regularized GRPO~\eqref{eq:grpo-loss} to planner tokens only.
\EndFor
\Statex
\Statex \textbf{Stage 4: Audit} (no parameter update)
\State Measure frozen-solver pass@$K$ and the fraction of rollouts with nonzero $\Rplan$; if either is insufficient, return to Stage~1 or 2.
\Statex
\Statex \textbf{Stage 5: Joint CPPO} \hfill\Comment{\S\ref{sec:loss}}
\For{$t = 1, \ldots, T_{\mathrm{cppo}}$}
  \ForAll{prompts $x$ in the batch}
    \State Sample $M$ planner tuples $S^{(m)} \sim q_\Theta(\cdot \mid x)$; score $\Jpsi(x, S^{(m)})$.
    \State For each $s_i^{(m)}$, sample $y_i^{(m)} \sim p_\Theta(\cdot \mid x, s_i^{(m)})$; verify $r_i^{(m)} = V(x, y_i^{(m)})$.
    \State Within-tuple normalize $\{r_i^{(m)}\}$ $\to$ solver advantages $a_i^{(m)}$~\eqref{eq:solver-adv}.
    \State $\Rout^{(m)} \gets \ind\!\left\{\max_i r_i^{(m)} = 1\right\}$; \quad $\Rplan^{(m)} \gets \Jpsi(x, S^{(m)}) \cdot \Rout^{(m)}$.
    \State Across-tuple normalize $\{\Rplan^{(m)}\}$ $\to$ planner advantages $A_{\mathrm{plan}}^{(m)}$~\eqref{eq:planner-adv}.
  \EndFor
  \State Apply GRPO solver loss with $a_i^{(m)}$ and GRPO planner loss with $A_{\mathrm{plan}}^{(m)}$; update $\Theta$ via~\eqref{eq:cppo-loss}.
  \State Optionally refresh $\psi$ on a fresh batch of planner outputs labeled by the same judge (Stage~2, \S\ref{sec:rm-training}).
\EndFor
\State \Return $\Theta$
\end{algorithmic}
\end{algorithm*}

\section{RM-Guided Planner Warm-Up}
\label{app:warmup}

\paragraph{Why planner warm-up helps.}
The RM-guided warm-up is not intended to directly improve solver
accuracy. During this stage the solver is frozen and no verifier
outcome is used; the planner is optimized only against the validity
gate,
\begin{equation}
\Rwarm(x, S) = \Jpsi(x, S).
\label{eq:warmup-reward}
\end{equation}
Its role is to align the planner's on-policy distribution with the
frozen plan-validity gate before joint CPPO. This alignment matters
because the planner reward in CPPO is multiplicative,
\begin{equation}
\Rplan(x, S) = \Jpsi(x, S) \cdot \Rout(x, S),
\label{eq:warmup-rplan}
\end{equation}
so when the planner frequently samples tuples rejected by $\Jpsi$,
solver success does not yield planner credit: $\Rplan = 0$ even when
$\Rout = 1$. Warm-up reduces these invalid rollouts and raises the
density of non-zero gated outcome rewards, which stabilizes the
subsequent joint planner--solver update. Warm-up therefore provides
only a small standalone gain under a frozen solver, but it gives the
joint CPPO stage a denser reward signal.

\section{Decoded-Token Accounting}
\label{app:decoded-token-accounting}

Table~\ref{tab:decoded-token-accounting} gives the decoded-token
accounting for the main-text token-normalized slices. Planner and
solver tokens are separated where the run logs expose that split:
planner tokens count generated plans or strategy tuples, and solver
tokens count generated solutions. Prompt tokens are excluded here,
matching Table~\ref{tab:decoded-token-efficiency}. The planner--solver
split is unavailable for the LCBv6 PlanSearch and CPPO runs (the
harness recorded only aggregate decoded length); the corresponding
cells are dashed, but the total decoded-token counts and the
pass@$K$/10k values are complete.

\begin{table*}[!h]
\centering
\caption{Decoded-token accounting for Qwen3.5-4B pass@$4$ slices.
All values are averaged per problem except pass@$K$ and
pass@$K$/10k tokens.}
\label{tab:decoded-token-accounting}
\small
\setlength{\tabcolsep}{3pt}
\begin{tabular*}{\textwidth}{@{\extracolsep{\fill}}lllrrrrrr@{}}
\toprule
Model & Dataset & Method & $K$ & Planner toks. & Solver toks. & Total decoded toks. & pass@$K$ & pass@$K$/10k toks. \\
\midrule
Qwen3.5-4B & APPS & Direct Solve & 4 & 0.0 & 6086.8 & 6086.8 & 0.515 & 0.846 \\
Qwen3.5-4B & APPS & Plan-and-Solve & 4 & 0.0 & 7326.2 & 7326.2 & 0.530 & 0.723 \\
Qwen3.5-4B & APPS & PlanSearch & 4 & 566.0 & 4969.0 & 5535.0 & 0.554 & 1.001 \\
Qwen3.5-4B & APPS & \textbf{CPPO} & 4 & \textbf{131.1} & \textbf{4596.7} & \textbf{4727.8} & \textbf{0.784} & \textbf{1.658} \\
\midrule
Qwen3.5-4B & LCBv6 & Direct Solve & 4 & 0.0 & 1731.0 & 1731.0 & 0.214 & 1.236 \\
Qwen3.5-4B & LCBv6 & Plan-and-Solve & 4 & 0.0 & 3975.0 & 3975.0 & 0.255 & 0.642 \\
Qwen3.5-4B & LCBv6 & PlanSearch & 4 & -- & -- & 2137.0 & 0.236 & 1.104 \\
Qwen3.5-4B & LCBv6 & PKPO & 4 & 0.0 & 2137.0 & 2137.0 & 0.488 & 2.284 \\
Qwen3.5-4B & LCBv6 & UpSkill & 4 & 0.0 & 2218.0 & 2218.0 & 0.411 & 1.853 \\
Qwen3.5-4B & LCBv6 & \textbf{CPPO} & 4 & -- & -- & \textbf{1556.0} & \textbf{0.505} & \textbf{3.246} \\
\bottomrule
\end{tabular*}
\end{table*}

\section{LiveCodeBench-v6 Pass@$K$ Sweep}
\label{app:lcb-sweep}

\begin{figure*}[!h]
\centering
\begin{subfigure}[b]{0.48\linewidth}
\centering
\includegraphics[width=\linewidth]{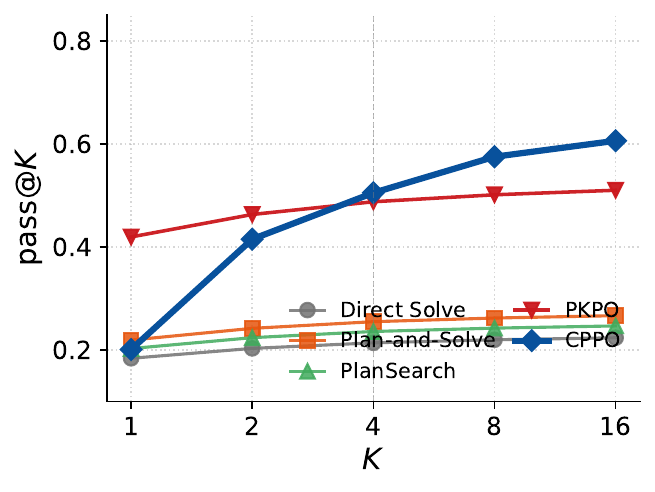}
\caption{Qwen3.5-4B}
\label{fig:passk-delta-full-4b}
\end{subfigure}\hfill
\begin{subfigure}[b]{0.48\linewidth}
\centering
\includegraphics[width=\linewidth]{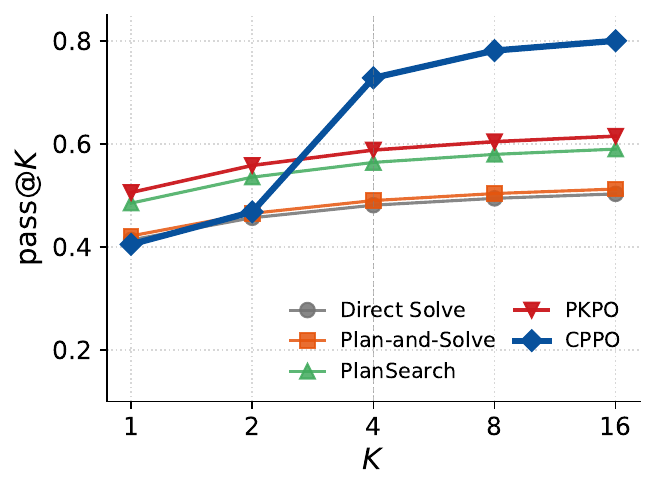}
\caption{Qwen3.5-9B}
\label{fig:passk-delta-full-9b}
\end{subfigure}
\caption{LiveCodeBench-v6 pass@$K$ across baselines for Qwen3.5-4B and
Qwen3.5-9B.}
\label{fig:passk-delta-full}
\end{figure*}

Figure~\ref{fig:passk-delta-full} extends Figure~\ref{fig:passk-delta} of
the intro with the Qwen3.5-4B panel alongside the Qwen3.5-9B panel
already shown in the intro. Both panels share the same $K \in \{1, 2,
4, 8, 16\}$ sweep on LiveCodeBench-v6. CPPO follows an S-shape on both
sizes: at $K{=}1$ the planner overhead leaves it at-or-below the
independent-sampling baselines; coordination starts to pay off from
$K{\geq}2$, the curve overtakes PKPO (the strongest non-CPPO baseline at
$K{=}4$ on both sizes) by $K{=}4$, and saturates from $K{=}8$ onward.
The inflection sits earlier on the weaker 4B base ($K{=}1{\to}2$) than
on the 9B ($K{=}2{\to}4$): a stronger solver makes a single strategy
more often sufficient at small budgets, delaying the point at which a
second alternative starts to pay off.

\section{Generality across Base Models: Gemma~4}
\label{app:gemma}

The main results in Table~\ref{tab:main} use the Qwen3.5 family
throughout. To check that CPPO's gains are not tied to a specific
backbone, we rerun the full pipeline on Gemma~4 E2B and
E4B~\citep{google2026gemma4} without retuning any component. The
training data follows the Qwen3.5 protocol: CodeContests supplies
planner SFT, reward-model data, planner warm-up, and joint CPPO
rollouts; APPS and LiveCodeBench-v6 remain evaluation-only. The
planner/solver roles, RM warm-up, GRPO hyperparameters, $K{=}4$
budget, verifier, and evaluation protocol are otherwise
identical to the Qwen3.5 runs in \S\ref{sec:evaluation}; we only swap in
Gemma~4 base weights for both
planner and solver. Table~\ref{tab:gemma} reports pass@$4$ on APPS and
LiveCodeBench-v6. CPPO achieves the highest pass@$4$ at both sizes on
both benchmarks, matching the pattern from the Qwen3.5 sweep in
Table~\ref{tab:main}. The Gemma~4 base is already much stronger on
APPS than Qwen3.5 of comparable size (Direct Solve $0.640/0.750$ on
E2B/E4B versus $0.420/0.515$ on Qwen3.5-2B/4B), so the absolute CPPO
margin over Direct Solve compresses ($+0.108/+0.082$ on E2B/E4B,
versus $+0.126/+0.279$ on Qwen3.5-2B/4B) -- the dominant headroom on
APPS is in the strongest baseline rather than the base model. The
LCBv6 pattern matches the Qwen3.5-2B sweep more directly.

\begin{table*}[h]
\centering
\caption{Pass@$4$ on APPS and LiveCodeBench-v6 for Gemma~4 E2B and
E4B~\citep{google2026gemma4}. LCBv6 Direct Solve cells are taken from
the released Gemma~4 LCBv6 evaluations~\citep{google2026gemma4}; all
other cells are trained and evaluated under the protocol of
\S\ref{sec:evaluation} at the same $K{=}4$ solver-attempt budget.}
\label{tab:gemma}
\scriptsize
\setlength{\tabcolsep}{3pt}
\renewcommand{\arraystretch}{0.85}
\begin{tabular}{@{}lcccc@{}}
\toprule
& \multicolumn{2}{c}{APPS} & \multicolumn{2}{c}{LiveCodeBench-v6} \\
\cmidrule(lr){2-3} \cmidrule(lr){4-5}
Method & E2B & E4B & E2B & E4B \\
\midrule
Direct Solve              & 0.640\std{0.019} & 0.750\std{0.018} & 0.440$^{\dagger}$ & 0.520$^{\dagger}$ \\
Plan-and-Solve            & 0.660\std{0.020} & 0.770\std{0.019} & 0.452\std{0.014} & 0.531\std{0.017} \\
PlanSearch          & 0.690\std{0.022} & 0.790\std{0.020} & 0.481\std{0.014} & 0.560\std{0.017} \\
Pass@$K$ Training / RLVR  & 0.708\std{0.018} & 0.798\std{0.015} & 0.519\std{0.019} & 0.598\std{0.015} \\
PKPO                      & 0.725\std{0.020} & 0.815\std{0.017} & 0.521\std{0.017} & 0.601\std{0.015} \\
\textbf{CPPO}             & \textbf{0.748}\std{0.021} & \textbf{0.832}\std{0.018} & \textbf{0.561}\std{0.019} & \textbf{0.629}\std{0.020} \\
\bottomrule
\end{tabular}

\smallskip
{\scriptsize $^{\dagger}$Single-point estimate from~\citep{google2026gemma4}; seed variance not reported.}
\end{table*}

\section{APPS Pass@$K$ Sweep}
\label{app:passk-sweep}

\begin{figure*}[!h]
\centering
\includegraphics[width=\linewidth]{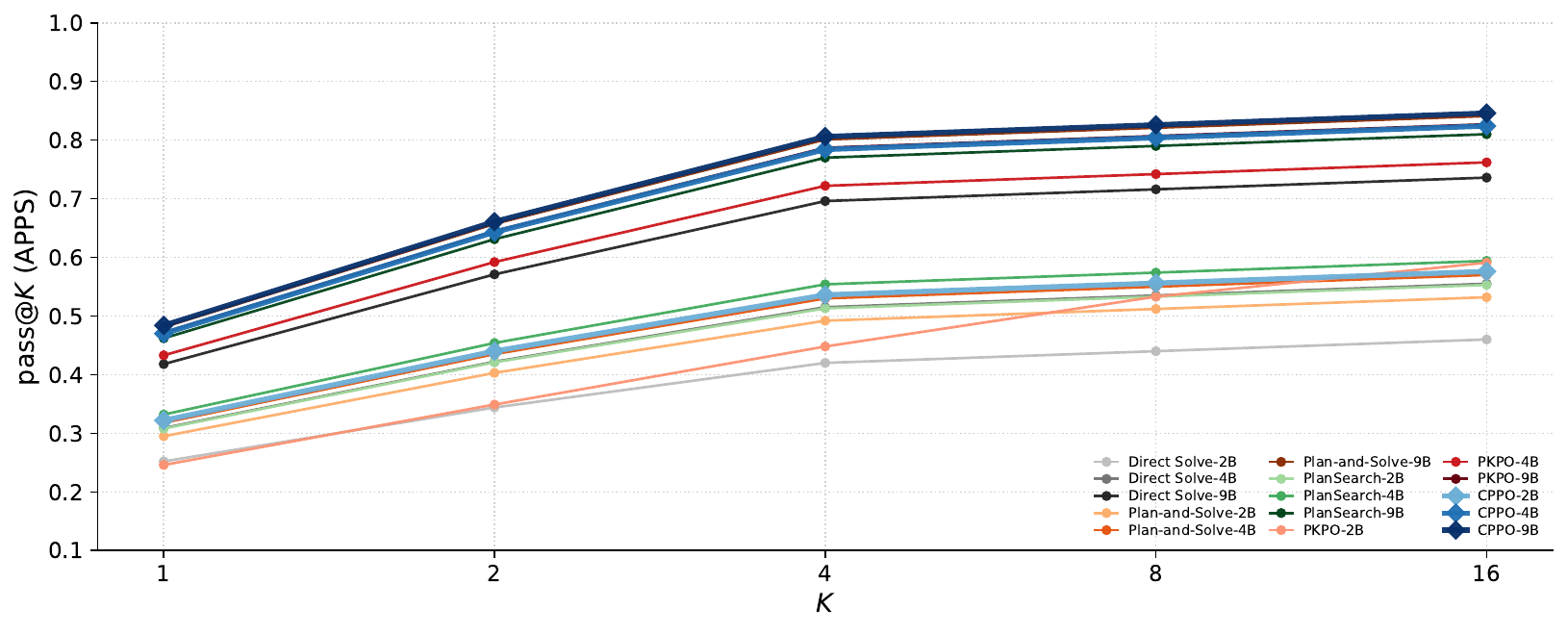}
\caption{APPS pass@$K$ across Qwen3.5-2B, Qwen3.5-4B, and Qwen3.5-9B. Each
method occupies a single color, with darker shades for larger models.}
\label{fig:passk-curve}
\end{figure*}

\begin{figure*}[!h]
\centering
\includegraphics[width=\linewidth]{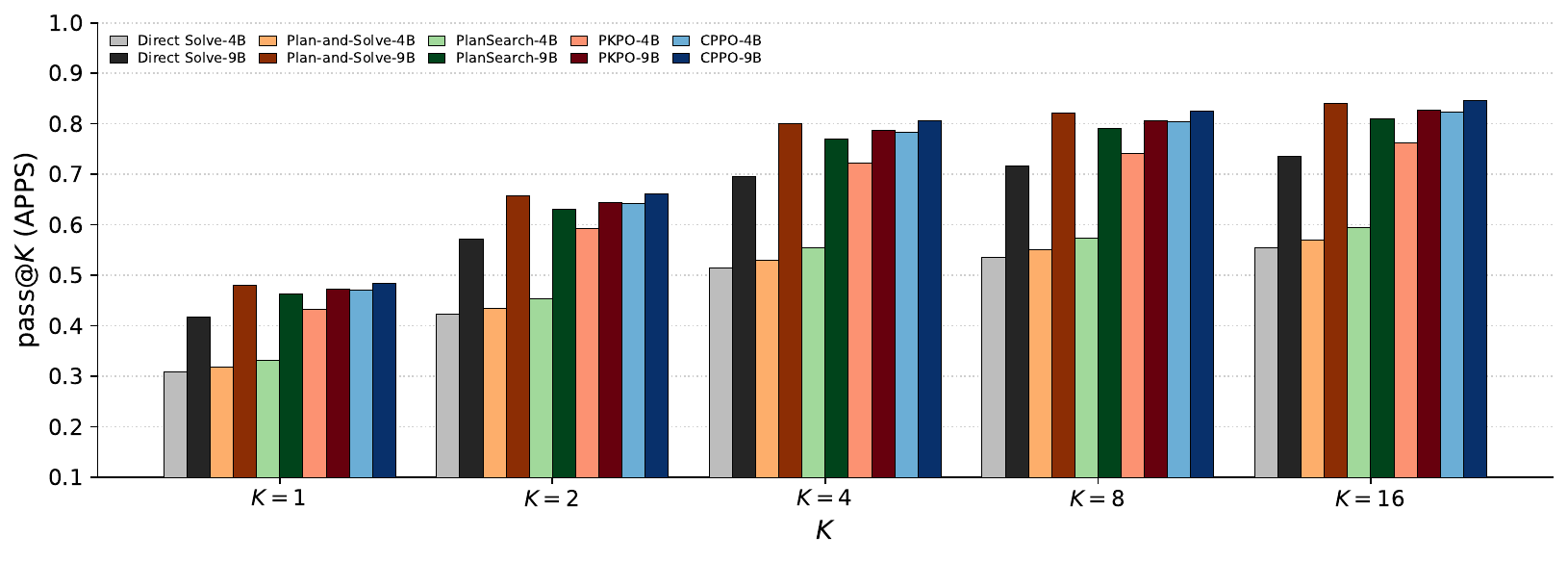}
\caption{Grouped APPS pass@$K$ for Qwen3.5-4B and Qwen3.5-9B at
$K \in \{1, 2, 4, 8, 16\}$. Each method occupies a single color across both
sizes, with a lighter shade for 4B and a darker shade for 9B.}
\label{fig:passk-bars}
\end{figure*}

Figure~\ref{fig:passk-curve} shows the full APPS pass@$K$ line sweep
for all baselines across Qwen3.5-\{2B, 4B, 9B\}; Figure~\ref{fig:passk-bars}
gives the same data as a grouped-bar view at $K \in \{1, 2, 4, 8, 16\}$
for the 4B and 9B sizes.

\paragraph{Coordination unit and per-$\Ksolve$ protocol.}
CPPO's coordination unit is $\Ktuple$: one planner rollout emits a
single coordinated tuple of size $\Ktuple$, and the solver
attempts each strategy in the tuple. The Figure~\ref{fig:passk-delta}
sweep uses one CPPO planner trained at $\Ktuple{=}4$ throughout (the
same checkpoint as Table~\ref{tab:main}, with no $K$-specific
retraining). For $\Ksolve \leq 4$ we take the first $\Ksolve$ outputs
of one $4$-tuple rollout; for $\Ksolve > 4$ we pool
$\lceil \Ksolve / 4 \rceil$ independent $4$-tuple rollouts. The
solver outputs are pooled before computing pass@$\Ksolve$. Coordination
is preserved \emph{within} each tuple, not across them: for
$\Ksolve > 4$, pass@$\Ksolve$ measures pooled coordinated tuples
versus iid samples at matched $\Ksolve$, not a single $\Ksolve$-way
joint coordination. Table~\ref{tab:k-sweep-decomposition} gives the
decomposition for each $\Ksolve$ on the sweep of
Figure~\ref{fig:passk-delta}.

Every method receives the same $\Ksolve$ solver-attempt budget. Direct
Solve draws $\Ksolve$ independent base-model samples. Plan-and-Solve
and PlanSearch draw $\Ksolve$ samples through each method's
plan-selection procedure. PKPO and Pass@$K$-only RL sample $\Ksolve$
outputs from their trained policies. CPPO additionally consumes
planner tokens at each tuple rollout; decoded-token accounting is
reported in Appendix~\ref{app:decoded-token-accounting}.

\begin{table}[!h]
\centering
\caption{CPPO rollout composition for the pass@$K$ sweep of
Figure~\ref{fig:passk-delta}. The same $\Ktuple{=}4$ planner is used
at every $\Ksolve$: rows with $\Ksolve < 4$ truncate one $4$-tuple
to its first $\Ksolve$ outputs; rows with $\Ksolve \geq 4$ pool
$\lceil \Ksolve / 4 \rceil$ independent $4$-tuple rollouts.
Coordination is preserved within each $4$-tuple, not across tuples.}
\label{tab:k-sweep-decomposition}
\small
\setlength{\tabcolsep}{10pt}
\renewcommand{\arraystretch}{1.05}
\begin{tabular}{cl}
\toprule
$\Ksolve$ & CPPO rollout composition \\
\midrule
$1$  & first $1$ of one $4$-tuple \\
$2$  & first $2$ of one $4$-tuple \\
$4$  & one $4$-tuple \\
$8$  & two $4$-tuples ($4{+}4$) \\
$16$ & four $4$-tuples ($4{+}4{+}4{+}4$) \\
$20$ & five $4$-tuples ($4{\times}5$) \\
$24$ & six $4$-tuples ($4{\times}6$) \\
\bottomrule
\end{tabular}
\end{table}

\section{Practical Rollout Recipe}
\label{app:rollout-recipe}

A minimal CPPO run proceeds in a short pipeline, each stage building on the
artifacts produced by the previous one.

We begin by assembling the planner-candidate pool: we sample from the current
planner checkpoint under the same prompt that CPPO will later use. An offline
LLM judge~\citep{zheng2023judge} labels every candidate against a structured
plan-validity rubric, and we balance the resulting train and validation
splits by prompt and by binary label.

We then train a small planner reward model on this dataset with binary
cross-entropy on the pass/fail target. Warm-up follows: holding the reward
model frozen, we update only planner tokens under $\Rwarm = \Jpsi(x, S)$ and
stop when the reward-model pass rate plateaus. We integrate the frozen reward
model into the planner--solver environment and run a brief smoke test under
$\Rplan = \Jpsi \cdot \Rout$, confirming that rewards are nonzero before we commit
to a larger run.

The main run launches from this checkpoint and tracks the full diagnostic
panel: outcome reward, plan-validity reward, the multiplicative planner reward,
reward sparsity, and planner and solver quality, so that drift in any of
them is caught early.

Because the planner's distribution shifts as training proceeds, we
periodically refresh the reward-model dataset: we label outputs from the
updated planner with the same judge, append them to the training pool,
rebalance by prompt and pass/fail label, and resume finetuning the reward
model from its latest checkpoint before CPPO continues.

\section{Baseline Implementation Details}
\label{app:baselines}

This section details how the baselines reported in Table~\ref{tab:main}
are implemented. All methods use the same code-extraction, sandboxed
execution, verifier, and pass@$K$ definition, and all are evaluated under
the same $K$-attempt budget as CPPO. The inference-only baselines keep the
base-model solver frozen; the trained baselines, PKPO and UpSkill, update
the policy and then emit $K$ answer samples at test time.

\paragraph{Training and selection protocol for trained baselines.}
PKPO and UpSkill use the same Qwen3.5 model sizes and the same CodeContests
train pool as the CPPO rollout stage. Their optimization recipes are fixed
on the CodeContests-train dev subset listed in
Table~\ref{tab:exposure-audit}, with no APPS, CodeContests-valid, or LCBv6
held-out examples used for hyperparameter choice, checkpoint selection, or
early stopping. To keep the comparison controlled, we reuse the CPPO/PKPO
optimizer and sampling settings where applicable: fp32 trainable weights,
AdamW learning rate $5{\times}10^{-7}$, clip ratio $\epsilon{=}0.2$, KL
coefficient $0.01$, temperature $0.7$, and maximum $2048$ decoded tokens.
Method-specific knobs are set to the paper defaults or to the pass@$4$
budget: PKPO uses $n{=}16$ and $k_{\mathrm{opt}}{=}4$, while UpSkill uses
$n_{\mathrm{skills}}{=}4$ and $\alpha_{\mathrm{mi}}{=}5.0$.

\paragraph{Plan-and-Solve \citep{wang2023plan} as a prompting baseline.}
\begin{enumerate}[leftmargin=*,itemsep=2pt,topsep=2pt]
\item The same base-model solver is given a prompt that asks it to first write
a solution plan and then write the code.
\item For each problem we sample $K$ full solution attempts under that prompt.
\item Each attempt passes through the same code extraction, sandboxed
execution, and verifier pipeline.
\item Pass@$K$ is the fraction of problems on which at least one of the $K$
attempts passes the tests.
\end{enumerate}
This comparison isolates the effect of adding a planning prompt before code
generation.

\paragraph{PlanSearch \citep{wang2024plansearch} as a planning baseline.}
We follow the released implementation, using eight candidate plans per problem
and the same solver/verifier pipeline as the other baselines.
\begin{enumerate}[leftmargin=*,itemsep=2pt,topsep=2pt]
\item The base model first generates multiple candidate plans per problem;
we use $8$ candidates.
\item The candidates are selected and organized into plans usable for solving.
\item The same frozen base-model solver then generates code conditioned on each
selected plan.
\item Each problem is evaluated under a budget of $K$ solver attempts, with the same verifier and pass@$K$ definition as above.
\end{enumerate}
This comparison isolates inference-time multi-plan search without planner
training.

\paragraph{PKPO \citep{walder2025pkpo} as a pass@$K$ RL baseline.}
PKPO transforms the per-sample reward vector $r \in \{0,1\}^n$ over $n \geq k$
samples of the same problem into a low-variance unbiased estimator of the
pass@$k$ gradient. We re-implement it on Qwen3.5-\{2B, 4B, 9B\} for code
generation (the original paper uses Gemma-2 on math): each gradient step
samples $n{=}16$ completions per problem, scores them with the same sandboxed
verifier, and applies PKPO's transform with $k_{\mathrm{opt}}{=}4$ to obtain
the REINFORCE-style update; no planner and no strategy tuple. At test time
the trained policy emits $K{=}4$ independent answer samples and is evaluated
under the same pass@$K$ verifier as CPPO. Training dynamics for the three
PKPO runs are shown in Appendix~\ref{app:pkpo-training}; we use them as a
sanity check that the reimplementation behaves as PKPO intends
(variance-reduced, advantage-sparsifying updates) before treating it as a
baseline.

\paragraph{UpSkill \citep{shah2026upskill} as a diversity-oriented RL baseline.}
UpSkill is also a trained baseline rather than a frozen-solver prompt. It
keeps the independent-answer policy class but conditions each sample on a
latent skill prefix; at $K{=}4$, we draw one completion from each of four
skill prefixes and score the resulting set with the same verifier. We adapt
UpSkill to the same full-finetuning setting as PKPO and CPPO: the original
rank-$32$ LoRA adapter is replaced by full-parameter training, while the MI
weight $\alpha_{\mathrm{mi}}{=}5.0$ and the remaining optimizer settings
follow the protocol above. As with PKPO, the training dynamics in
Appendix~\ref{app:upskill-training} confirm that the full-finetuning
adaptation preserves UpSkill's intended behavior---a mutual-information
reward that rises alongside the correctness reward---before we use it as a
baseline.

\section{PKPO Training Dynamics}
\label{app:pkpo-training}

\begin{figure*}[!h]
\centering
\includegraphics[width=\linewidth]{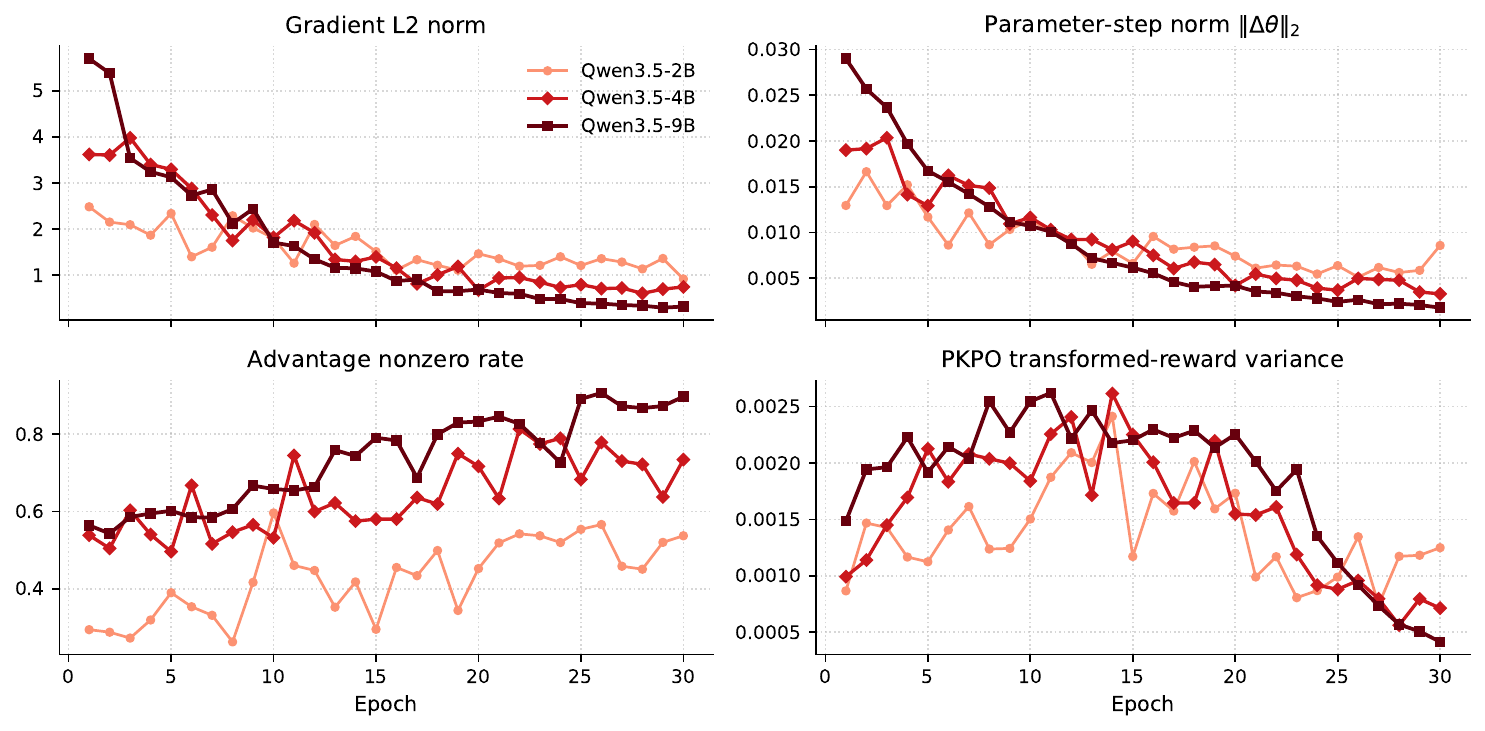}
\caption{PKPO training dynamics for Qwen3.5-2B, 4B, and 9B over 30 epochs
(light to dark red): gradient L2 norm, parameter-step norm
$\|\Delta\theta\|_2$, advantage nonzero rate after group normalization,
and variance of the PKPO-transformed reward.}
\label{fig:pkpo-training}
\end{figure*}

Figure~\ref{fig:pkpo-training} tracks the PKPO update across 30 epochs
for Qwen3.5-2B, 4B, and 9B, trained on the same CodeContests train pool
as our other RL methods (see Table~\ref{tab:data-splits}). Each step
draws $n{=}16$ completions per problem at temperature $0.7$, applies the
SLOO-minus-one transform of \citet{walder2025pkpo} with
$k_{\mathrm{opt}}{=}4$, and updates with AdamW at learning rate
$5{\times}10^{-7}$, clip ratio $\epsilon{=}0.2$, and KL coefficient
$0.01$ against a frozen reference. The three sizes share these settings;
only the base checkpoint differs.

The gradient and parameter-step norms decay as the policy converges, with
9B contracting fastest and 2B plateauing highest---larger models spread
the same outcome signal over more parameters. The advantage nonzero rate
is the fraction of problems still producing gradient: the SLOO-minus-one
transform zeros it out whenever a problem is entirely solved or entirely
failed, and larger models cross out of the all-fail regime sooner
($\approx\!0.55$ at 2B versus $\approx\!0.85$ at 9B). The
transformed-reward variance rises early as the policy first generates
mixed pass/fail groups, plateaus through mid-training, and falls late as
the policy converges on the solvable subset; the late drop is sharpest
at 9B and most muted at 2B.

The 2B trace also matches the PKPO 2B rows of
Table~\ref{tab:main-full}, where PKPO trails Tuple Planner SFT on both
APPS and CodeContests: the elevated gradient plateau, the lowest
advantage-nonzero rate of the three sizes, and the muted late variance
drop together describe a sparse-signal regime in which most groups
remain entirely failed and the SLOO-minus-one update has little to
work with.

\section{UpSkill Training Dynamics}
\label{app:upskill-training}

UpSkill shares PKPO's policy class but adds a token-level
mutual-information reward $r_{\mathrm{mi}}$ to the binary correctness
reward $r_{\mathrm{correct}}$. The MI term sets UpSkill apart: PKPO's
signal vanishes whenever a problem is entirely solved or entirely failed,
while UpSkill's MI term continues to produce a nonzero per-token gradient
as long as the latent skill code distinguishes trajectories. The
diagnostic panels therefore differ from Figure~\ref{fig:pkpo-training}:
we report the gradient L2 norm, the two reward components, and the
combined-reward variance.

Each step draws one completion per skill at temperature $0.7$ under a
fixed prompt prefix encoding the latent skill, with $n_{\mathrm{skills}}{=}4$
(matched to the $K{=}4$ inference budget) and MI weight
$\alpha_{\mathrm{mi}}{=}5.0$. Optimizer settings mirror PKPO above; we
replace the rank-$32$ LoRA adapter of the original UpSkill with full
fine-tuning across all three sizes, leaving the rest of the
hyperparameters unchanged. Parameter-step norm is omitted because the
training logger did not record it.

The gradient L2 norm decays for every size, with the 9B trace smoother
and the 2B trace visibly noisier. The correctness reward
$r_{\mathrm{correct}}$ rises along a sigmoid whose asymptote scales with
model size, consistent with stronger models reaching higher pass@$4$. The
MI reward $r_{\mathrm{mi}}$ saturates early and stays high across all
sizes: the latent-skill conditioning keeps trajectories distinguishable
even after correctness plateaus, so the MI term can continue to supply
token-level signal after correctness rewards become sparse. The combined-reward variance falls over
training as the policy concentrates on a smaller set of higher-quality
skills.

\begin{figure*}[!h]
\centering
\includegraphics[width=\linewidth]{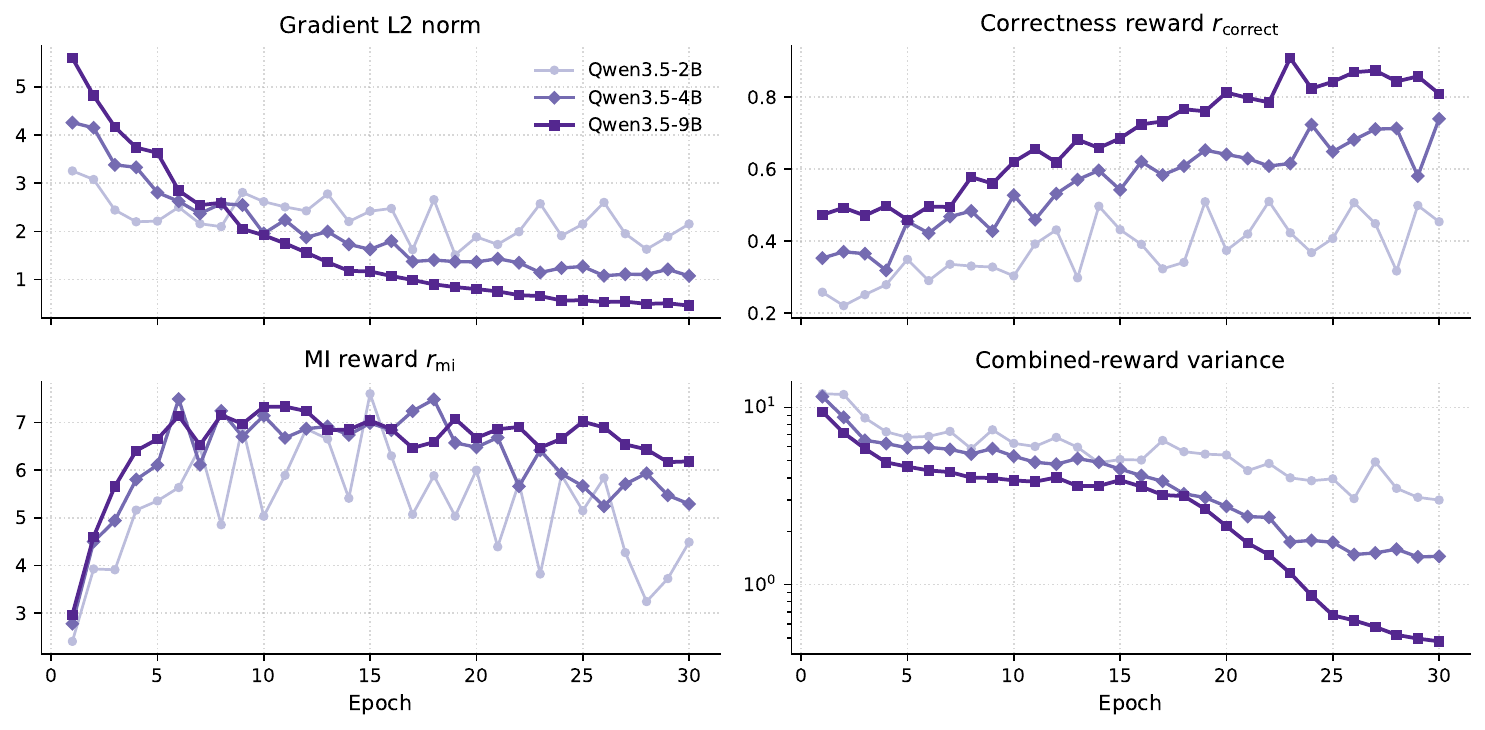}
\caption{UpSkill training dynamics for Qwen3.5-2B, 4B, and 9B over 30
epochs (light to dark purple): gradient L2 norm, correctness reward
$r_{\mathrm{correct}}$, mutual-information reward $r_{\mathrm{mi}}$, and
combined-reward variance.}
\label{fig:upskill-training}
\end{figure*}

\end{document}